%% file: T4D_FaR.tex
\documentclass{article}
\usepackage[table,xcdraw]{xcolor}
\usepackage{iclr2024_conference}
\iclrfinalcopy

\input{math_commands.tex}

\usepackage{times}
\usepackage{latexsym}

\usepackage[T1]{fontenc}
\usepackage[utf8]{inputenc}

\usepackage{float}
\usepackage[hidelinks]{hyperref}
\usepackage{verbatim}
\usepackage{wrapfig}
\usepackage{lipsum}
\usepackage{amssymb}
\usepackage{amsmath}
\usepackage{booktabs}
\usepackage{url} 
\usepackage[inline]{enumitem}
\usepackage{mathtools}
\usepackage{subcaption}
\usepackage{algorithm}
\usepackage{algorithmic}
\usepackage{fontawesome} 
\usepackage{contour}
\usepackage{textcomp}
\usepackage{amsmath}

\usepackage{microtype}
\usepackage{xspace}

\usepackage[symbol]{footmisc}

\usepackage{dsfont}
\definecolor{antiquefuchsia}{rgb}{0.57,0.36, 0.51}
\definecolor{teal}{RGB}{0, 128, 128}

\usepackage{microtype}
\usepackage{inconsolata}

 \newcommand{\task}{\textsc{T4D}\xspace}
\newcommand{\frameworklong}{\textbf{F}oresee \textbf{a}nd \textbf{R}eflect}
\newcommand{\framework}{\textsc{FaR}\xspace}

\title{How FaR Are Large Language Models From Agents with Theory-of-Mind?}
\author{Pei Zhou$^{\diamondsuit}$\thanks{Work done during Google Internship} \quad
Aman Madaan$^{\spadesuit}$ \quad
Srividya Pranavi Potharaju $^{\dagger}$ \quad
Aditya Gupta \quad \\
\textbf{Kevin R. McKee}$^{\ddagger}$ \quad
\textbf{Ari Holtzman}$^{\clubsuit}$ \quad
\textbf{Jay Pujara}$^{\diamondsuit}$ \quad
\textbf{Xiang Ren}$^{\diamondsuit}$ \quad
\\
\textbf{Swaroop Mishra}$^{\ddagger}$ \quad
\textbf{Aida Nematzadeh}$^{\ddagger}$ \quad
\textbf{Shyam Upadhyay}$^{\dagger}$ \quad
\textbf{Manaal Faruqui}$^{\dagger}$ \quad
\\
\small{$\dagger$ Google} \quad
\small{$\ddagger$ Google DeepMind} \quad
\small{$\diamondsuit$ University of Southern California}  \\
\small{$\spadesuit$ Carnegie Mellon University} \quad 
\small{$\clubsuit$ University of Chicago} \quad
\\\small \centering {\texttt{\href{mailto:peiz@usc.edu}{peiz@usc.edu}}}
}

\begin{document}
\maketitle
% \floatsetup[table]{capposition=top}

\begin{figure*}[h]
	\centering
    \vspace{-0.3cm}
	\includegraphics[width=\columnwidth]{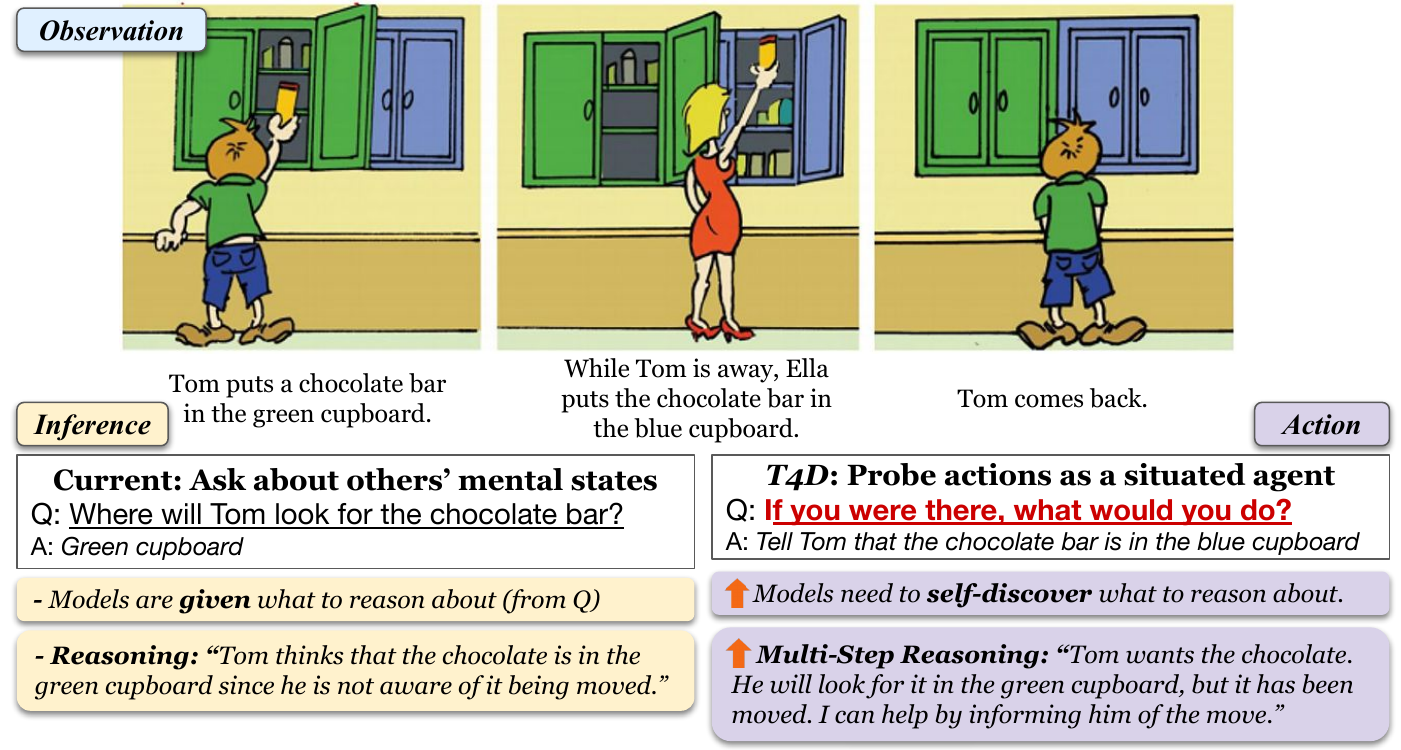}
% 	\vspace{-0.1cm}
	\caption{%\vspace{-0.5cm} 
    % \small
	Given \textit{observations}, current social reasoning tasks ask models questions targeting specific \textit{inferences} (left). We propose T4D to probe whether LLMs can decide proper \textit{actions} using theory-of-mind as a situated agent (right). They key challenges in T4D are 1) models have to \textit{identify} relevant inferences about mental states without being directed towards one and 2) to arrive at proper action choices, more steps of reasoning are required.}
 % \swaroop{The figure is still ambiguous, also the caption needs to be improved for example: what does the yellow block of text shows, the purple one shows etc. Caption needs to be detailed. Also not every reader is aware of the Tomi dataset, you may want to explain a bit more since this image is the very first thing reader will see} \pei{updated}
	  	 % \vspace{-0.5cm}
	\label{fig:motivation}
\end{figure*}
\begin{abstract}
\input{sections/0_abstract.tex}
\end{abstract}

\section{Introduction}\label{intro}

\input{sections/1_intro.tex}

\section{Background and Related Work}\label{rel_work}

\input{sections/2_background}

\section{Thinking for Doing (T4D): Task and Data}\label{task}

\input{sections/3_formulation}

\section{LLMs Struggle on T4D While Humans Find it Easy}\label{T4D_results}

\input{sections/4_T4D_perf}

\section{Foresee and Reflect (FaR) Prompting}\label{FaR}
\input{sections/5_FaR}

\section{FaR Boosts LLM Dramatically and Generalizes Robustly}\label{FaR_results}
\input{sections/6_experiments}

\newpage
\section{Conclusion}\label{conclusion}
\input{sections/7_conclusion}

% \section{Limitations}
% \input{sections/8_limitations}\label{limitations}

\bibliography{anthology, custom,iclr2024_conference}
\bibliographystyle{iclr2024_conference}

%\appendix
\clearpage
\appendix
\label{appendix}
\input{sections/9_appendix.tex}

\end{document}

%% file: math_commands.tex
\usepackage{amsmath,amsfonts,bm}

\def\eqref#1{equation~\ref{#1}}
\def\1{\bm{1}}

\DeclareMathAlphabet{\mathsfit}{\encodingdefault}{\sfdefault}{m}{sl}
\SetMathAlphabet{\mathsfit}{bold}{\encodingdefault}{\sfdefault}{bx}{n}

%% file: sections/0_abstract.tex
% w/ Jay: Baseline: append ToMi questions and answers before answering T4D
% Benevolence assumption?
% other keywords: personalization? robustness
% what is modeling benefit; not just task benefit; what would LLMs get from this
% --> AI systems intervene in some way, current benchmarks not enough, we want models to enact helpful changes to the world

% act, does not have to always help;
% over 2 T4D domains, provide information and emotional support

% \swaroop{The theme of the title 'X <verb> LLMs to <task>' is awesome; it resembles with CoT paper title so will get easy attention. However the current X is not great here specifically the word heuristics does not sound good, may be "future thinking/anticipation" or something like that?}
\textit{Thinking is for Doing.}
Humans can \textit{infer} other people's mental states from observations--an ability called Theory-of-Mind (ToM)--and subsequently \textit{act} pragmatically on those inferences.
Existing question answering benchmarks such as ToMi ask models questions to make \textit{inferences} about beliefs of characters in a story, but do not test whether models can then use these inferences to guide their \textit{actions}.
We propose a new evaluation paradigm for large language models (LLMs): \textit{Thinking for Doing (T4D)}, which requires models to connect inferences about others' mental states to actions in social scenarios. 
% Specifically, we convert ToMi to T4D where models must \textit{infer} beliefs of story characters to choose one to \textit{help} due to them holding false beliefs.
Experiments on T4D demonstrate that LLMs such as GPT-4 and PaLM 2 seemingly excel at tracking characters' beliefs in stories, but they struggle to translate this capability into strategic action.
% Despite human-level performance on questions about others' beliefs, LLMs including GPT-4 and PaLM 2 show significant gap in performance when asked to choose an action in T4D, while humans show 95\% agreement on the correct action.
% \mfar{isn't this the same as CoT?} \mfar{Let's either stick to FaR of T4D, having both terms in title and abstract is a little much.} \pei{T4D is an evaluation paradigm, FaR/CoT are prompting methods}

% \mfar{unclear what "act" means here} \pei{changed wording, see if it sounds clearer}
% We argue that it is crucial for language agents to not only able to arrive at plausible inferences given a situation, but also to know how to choose actions based on the inferences. 
% We find that existing LLMs including ChatGPT, GPT-4, and PaLM 2, despite close-to-human performance on answering inference questions, perform close-to-random when determining what is the right action to do, while humans reach over 90\% agreement on the right option.
% Next, we study the ambiguities in our task, which naturally rise with this more realistic setup, and find models (especially GPT-4) performance improve when we pose a very specific question. 
% By analyzing why T4D is challenging, we posit that the key bottleneck is locating the proper implicit inference steps that lead to action intents.
Our analysis reveals the core challenge for LLMs lies in identifying the implicit inferences about mental states without being explicitly asked about as in ToMi, that lead to choosing the correct action in T4D.
To bridge this gap, we introduce a zero-shot prompting framework, \textit{Foresee and Reflect} (FaR), which provides a reasoning \textit{structure} that encourages LLMs to anticipate future challenges and reason about potential actions.
% \mfar{is it the problem or solution? please clarify}. \pei{changed, this is the bottleneck we find that leads to us proposing a solution (next sentence)}
% design a zero-shot prompting framework that provides generative models a \textit{structure} to elicit reasoning.
% to extract signals from themselves, as models might not have implicit access to information they can explicitly generate.
% To address this, we design a zero-shot prompting framework that prompt LLMs to \textit{extract signals} from themselves to link inferences with actions, as we hypothesize that such information is much harder for generative models to implicitly access than explicitly generating it.
% Specifically, we introduce \textit{Foresee and Reflect} (FaR), which prompts LLMs to predict potential future challenges for characters and then reflect on possible actions to mitigate these challenges. 
FaR boosts GPT-4’s performance from 50\% to 71\% on T4D, outperforming other prompting methods such as Chain-of-Thought and Self-Ask.
% \swaroop{not sure if self-ask is the right baseline to compare, decomposed prompting, least to most would have been probably more appropriate}. \pei{will add more explanation in related work, tldr is that self-ask applies directly on zero-shot with only 1 inference call while others need multiple and need decomp examples}
Moreover, FaR generalizes to diverse out-of-distribution story structures and scenarios that also require ToM inferences to choose an action, consistently outperforming other methods including few-shot in-context learning.
% Additionally, FaR robustly generalizes to diverse story structures and scenarios including where the model must decide who to provide emotional support, whereas other methods, including few-shot prompting, do not generalize as well.

% Specifically, we propose \textit{Foresee and Reflect} (FaR) that prompts LLMs to first \textit{foresee} potential future challenges of characters and then \textit{reflect} on what actions could be performed now to alleviate those challenges.
% Extensive experimental results show that FaR helps boost GPT-4's performance by 50\%, significantly outperforming other prompting methods such as Chain-of-Thought (CoT) and Self-Ask.
% Furthermore, FaR robustly generalizes to diverse story structures and a new-domain T4D task where the models needs to choose who to provide emotional support, while other methods including few-shot CoT prompting generalizes poorly.

% In other words, our prompt motivates generative models to \textit{extract signals} from themselves to link inferences with actions, 

% "using generative models to extract signal from themselves". Essentially, make the argument that models don't have implicit access to information they can explicitly generate. 

% to improve LLM's ability to handle situations with ambiguities that can generalize to unseen social scenarios by decomposing the question to more specific directions and combine with searching algorithms to arrive at the most plausible actions to perform.

%% file: sections/1_intro.tex
Humans act with specific intentions, often grounded in reasoning about their environment and the mental states of others.
% Theory-of-Mind (ToM), the ability to reason about others' intentions, emotions, and beliefs~\citep{premack1978does, frith2003development}, serves a critical role in social interactions.
 For example, if Tom's friend Anne is looking for her backpack in the office, and Tom knows it is in the kitchen, Tom will intervene to help Anne by suggesting she check the kitchen.
% \amanx{Tom's action is based on knowing that}
This proactive action stems from Tom's understanding of three aspects: 1) Anne's goal of finding her backpack; 2) the knowledge of backpack being in the kitchen; and 3) Anne's belief of thinking the backpack is in the office.
Reasoning about Anne's mental states allows Tom to conclude that the mismatch between belief and knowledge prevents Anne from reaching her goal, and his intervention can help.
% two aspects: 1) Anne will benefit from knowing the location of her backpack and 2) Anne is not aware of its location, both requiring reasoning about Anne's mental states including her goals, beliefs, etc.
% This ability is referred as Theory-of-Mind (ToM), which serves a critical role in social interactions \citep{premack1978does, frith2003development}.
% \aman{Suggestion: 
Such capabilities to reason about and act on another individual's beliefs, intentions, and emotions are referred to as Theory-of-Mind (ToM), a critical element of human social interactions \citep{premack1978does, frith2003development}
The rise of large language models (LLMs) has prompted extensive research into their potential for \textit{Theory-of-Mind (ToM)} capabilities~\citep{sap-etal-2022-neural, kosinski2023theory, ullman2023large, shapira2023clever}. 
These investigations predominantly rely on established psychological tests, such as the False Belief Test~\citep{wimmer1983beliefs, baron1985does, perner1987three}. 
While existing benchmarks~\citep{nematzadeh2018evaluating, le2019revisiting} gauge LLMs' proficiency in \textit{inferring} mental states from scenarios (see Figure~\ref{fig:motivation} left), they often overlook an essential human capability: \textit{acting\footnote{We use ``\textit{acting}'' to refer to performing action in social scenarios like providing information to others.} on inferred mental states}. 
Simply put: humans often act based on inferred intentions and beliefs. 
In contrast, despite LLMs' performance in the False Belief Test, they often fail to infer what actions would be most useful in scenarios that humans would find trivial, a crucial consideration for the development of next-generation AI agents, from virtual assistants to embodied robots.
% }

% it has done the correct reasoning, as humans do, because it has not been provide a chain of reasoning
% In other words, it is still unknown that whether LLMs exhibit \textit{functional competence}~\citep{mahowald2023dissociating} that requires ToM.

% P3: T4D (where we probe acts)
We introduce a new evaluation paradigm: \textit{Thinking for Doing (T4D)}~\citep[see][]{fiske1992thinking} to probe whether models can \textit{determine proper actions} based on the mental states of others, rather than merely being able to answer questions about mental states.
% examine whether models can use ToM to come up with proper \textit{intents} to proactively help human users to prevent inconvenience. 
% \amanx{On a high-level, T4D treats models as situated agents who are given a sequence of observations and are asked to select the most suitable action from a list of candidates.}{
At its core, T4D envisions models as agents processing a series of observations to determine the most apt action from a set of options.
% }
% The observations consist of descriptions of multiple characters' movements and actions and hint at one of the characters holding a false belief of a location of an item, which is the correct answer.
Specifically, we adopt stories from a widely-used ToM benchmark: ToMi~\citep{le2019revisiting}, based on Sally-Anne False Belief Test~\citep{baron1985does} into observations in T4D.
% so that it requires mental state reasoning to infer that one character holds a false belief and thus the agent should act to help by providing information (Figure~\ref{fig:motivation}).
This integration ensures that models must utilize mental state reasoning, particularly when a character is identified to hold a false belief (as depicted in Figure~\ref{fig:motivation}).
% \amanx{The key novelty of T4D compared to ToMi is that instead of directly asking questions about inferences from mental state reasoning, we take one step further and asks models to choose an intent to act that is based on understanding others' mental states.}{
The crux of T4D's novelty, as visualized in Figure~\ref{fig:motivation}, lies in its objective: instead of merely eliciting inferences from mental state reasoning, it compels models to determine \textit{actions} based on the former.
% }

% P4: T4D realistic while challenging; and humans got good; Why? we do a factorization (QD, ToM, CSA)
% P4.5 (merge): the reason we did it is ... gives us insights into the FaR methods *structure*, other approaches fall short because they lack the structure
T4D presents a new zero-shot challenge for LLMs. We find the highest performance (GPT-4) capped at 50\% while human annotators reach over 95\% agreement.
% \mfar{maybe mention the dataset size}.
% On a human study with 20 annotators per question, we find that over 90\% of sampled T4D instances reach over 95\% agreement.
% LLM's performance, however, drops significantly when shifting the task from ToM inference questions to T4D, with the highest accuracy (GPT-4) being 50\%, dropped from 93\% on ToMi.
To gain deeper insights into the challenges LLMs encounter in T4D, we identify three reasoning patterns from human-written rationales: question decomposition, theory-of-mind inferences, and commonsense assumptions.
% \amanx{Then we make T4D \textit{easier} by providing models with the oracle reasoning steps of each pattern.}{
Then we test LLMs in \textit{oracle} settings, providing models with oracle reasoning steps based on the identified patterns.
% }
% \aman{The \textit{make easier} phrase is a bit confusing, so I've added a rewrite.}
% \amanx{As we will show in Section~\ref{sec4.2:ablations}, LLM's struggle with T4D can be largely attributed to the limitations of models to \textit{infer the right evidence} that leads to the correct action since as soon as we provide hints on what are relevant inferences, LLMs' performance get close to humans.}{
As demonstrated in Section~\ref{sec4.2:ablations}, the primary challenge LLMs face in T4D is pinpointing the correct evidence to inform their actions. When we provide models with specific hints about relevant inferences, their performance significantly improves, approaching human levels.
% }
% We find that the bottleneck lies in model's capability to \textit{navigate the latent inference space} given observations and infer the right evidence ToM to choose the correct action intent. 
% Since the questions in T4D do not \textit{suggest} models what to reason about, but give models the freedom to explore all potential inferences.

% P5: generalization results, contribution
% \amanx{To help LLMs navigate the inference space, we propose a new zero-shot prompting framework~\frameworklong~(FaR) that guides model's inferences using \textit{future heuristics}.}{
The clear potential of LLMs to perform T4D with proper guidance leads to the question: 
Can we develop a method that improves LLMs' T4D performance \textit{without} providing oracle hints but instead teaching models to better \textit{structure} their reasoning process?
% This evident potential of LLMs in performing T4D, when guided appropriately, underscores a key question: how can we systematically harness this capability? 
In response, we introduce a new zero-shot prompting framework \frameworklong~(FaR) that guides model's inferences by providing a reasoning \textit{structure} using future thinking.
% }
FaR has two components: Foresee, where it prompts the models to \textit{predict future events based on observations} and Reflect, where models \textit{reason on which action choice better helps the characters with potential challenges}. 
% Our intuition is to help LLMs to extract signals from themselves by providing a reasoning \textit{structure} to generate inferences that bridge between observations and action intents.
Comparison with prompting strategies including Chain-of-Thought~\cite{wei2022chain}, Tree-of-Thought~\citep{yao2023tree} (zero-shot), and Self-Ask~\citep{press2022measuring} shows that FaR improves LLM zero-shot performance by as much as 50\% while other methods do not display significant improvement.

% \pei{stress generalization tests more, separate paragraph+contribution list?} 
To explore FaR's strengths and limitations in more depth, we perform ablation studies aiming to answer two questions: \textit{are both foresight and reflection needed for improving LLMs} and \textit{what happens if we feed models noisy future predictions}?
We find that both components are crucial for tackling T4D and that LLMs are sensitive to noisy reasoning steps about the future in FaR, making how to help LLMs \textit{recover} from noisy foresight an intriguing future direction.
To examine whether FaR overfits on the ToMi-converted T4D task, we also conduct \textit{generalization study} by testing on out-of-distribution story structures and a non-False-Belief ToM task. 
We find that FaR shows consistent improvement across generalization tests, even outperforming few-shot prompting.
Our contributions are as follows:
% \pei{contribution list can be skipped if space is an issue}
\begin{enumerate}
    \item We propose \textit{Thinking for Doing}, a evaluation paradigm to challenge whether models can connect social reasoning to actions.
    \item We find LLMs struggle on T4D and our analysis indicates the key bottleneck is identifying implicit inference steps.
    \item We design~\frameworklong~(FaR), a zero-shot prompting framework that dramatically improves LLMs' performance on T4D. Analysis and generalization studies show that FaR robustness generalize to diverse contexts.
\end{enumerate}

% \srividya{might be good to mention which category of questions/tasks see the most improvement, like a contrastive study, could be put in results/exp section}

%% file: sections/2_background.tex
% \paragraph{Theory-of-Mind in Psychology}

\paragraph{Theory-of-Mind and Language Models}
Theory-of-mind has been studied extensively in psychology and cognitive science~\citep{premack1978does, baron1985does, frith2003development}, and clinical psychology tests such as False Belief Test~\citep{wimmer1983beliefs} (FBT) were developed to test ToM abilities in children.
More recently, as neural language models (LM) display impressive performance in many language understanding tasks, more studies aim to answer whether LMs exhibit ToM~\citep{sap-etal-2022-neural, kosinski2023theory, ullman2023large,shapira2023clever, sclar2023minding, trott2023large} using False Belief-templated story datasets such as ToM-bAbI~\citep{nematzadeh2018evaluating} and ToMi~\citep{le2019revisiting}.
Though stories cover limited range of interactions, other sources of ToM tests also face challenges, such as scalability due to costs of human-generated interactions~\citep{bara2021mindcraft} and noises in text-game environments~\citep{zhou-etal-2023-cast}.
This work focuses on False-Belief tests for ToM, the most studied subarea, and revisits the format of such tasks when testing LLMs.
Specifically, while probing work shows that LLMs display some degree of ToM but lack robustness~\citep{sap-etal-2022-neural, shapira-etal-2022-interactive}, we find that when asked FBT in a more realistic scenario, models fail even on the unperturbed tasks.

\paragraph{Large Language Models and Agents}
A line of recent work aims to build \textit{language agents}~\citep{andreas-2022-language, mahowald2023dissociating} that can perform ``\textit{actions}''.
Actions range from mimicking human social behavior~\citep{park2023generative}, completing tasks using websites~\citep{gur2023real}, and tool using~\citep{yao2023react, schick2023toolformer}.
Our work distinguishes from them by focusing on actions that require proper mental state modeling of other individuals (ToM), attributing the performance gap between answering inference questions only and choosing actions based on inferences, and designed a zero-shot prompt that improves models' capability that robustly generalizes.

\paragraph{Prompting Techniques for LLM}
% \aman{I think we can use some of the space here to preempt comparisons with existing prompting techniques: world model, tot}
% Numerous prompts have been proposed to augment reasoning capabilities of LLMs.
% \textit{Chain-of-Thought} (CoT)~\citep{wei2022chain} ...
% \textit{Tree-of-Thought (ToT)}~\citep{yao2023tree} (Basic Zero-Shot): ToT proposes to organize intermediate ``\textit{thoughts}'' in a tree and designs few-shot prompts to decompose, generate, and evaluate ``\textit{thoughts}'' tailored to each task. 
% \swaroop{this section appears a bit weak (a) there are several prompting methods we dont cite here such as least to most, decomposed prompting, help me think prompting etc. (b) we dont mention how our work is different from others, this is also critical to ensure no comparision with them needed. (c) some general prompting literature like 'reframing instructional prompts', the prompting survey needs to be cited} \pei{to be added}

Recent advancements in the area of LLMs have given rise to a plethora of few-shot~\citep{brown2020language} and instruction~\citep{mishra_reframing_2021} prompting techniques, including \textit{Chain-of-Thought} prompting~(CoT)~\citep{wei2022chain}, Least-to-most prompting~\citep{zhou2022least}, and search-based approaches like \textit{Tree-of-Thought (ToT)}~\citep{yao2023tree}, {Graph-of-Thought}~\citep{got1,got2}, and RAP~\citep{rap}.

However, the primary objective of our work is not to introduce a new prompting technique. Instead, we focus on the benefits of imposing a structured framework on the LLM's reasoning process, particularly in the context of Theory of Mind (ToM) tasks. Specifically, our analysis~(Section~\ref{sec4.2:ablations}) reveals essential elements of reasoning that can help LLM agents act (Foresee (F) and Reflect (R)), and we capture this in our proposed approach FaR.
Moreover, any prompting method that supports granular, multi-step reasoning and captures the Foreseeing and Reflecting steps is well-equipped to address the intricacies of ToM tasks.
% \aman{Pei, please take a look.} \pei{thanks, looks good}

%% file: sections/3_formulation.tex
% w/ jay: State space and Action space
% Structure: chain of inferences; some; CoT underspecify the structure of the latent space, trick is to provide structure (action, ToM, background commonsense) and use in Section 4

% question is task (T_I and T_A?)
% i should be before on the left side, in ablations we are adding on the right side
% base is only generating the answer; jointly generatinbg a rationale with an answer, for each of them, what is conditioned on and what is generated; compare inference space and with different structures; 

Here we formulate the \textit{Thinking for Doing (T4D)} task that requires models to use social reasoning to choose a proper action as a situated agent.

\subsection{T4D Task}\label{sec:3.1}

% Consider four variables in grounded social scenarios from an agent's perspective: \textit{observations} $\mathcal{O}$, a \textit{task} $\mathcal{T}$ verbalized by the question and answer options, \textit{inferences} $\mathcal{I}$ made by the agent based on the observations, and the \textit{action} $\mathcal{A}$ determined by the agent given observations and inferences. 
% For example, in Figure~\ref{fig:motivation}, $\mathcal{O}$ is shown as the 3-panel comic with captions, $\mathcal{T}$ is in the enclosed boxes with questions, $\mathcal{I}$ is shown as the answer in the left enclosed box, and action $\mathcal{A}$ for T4D is the answer in the right enclosed white box. 
% \pei{connecting variables to fig1 example is awkward, I'll add more annotations in fig1}
% \aman{Example of these four terms here
\newcommand{\obs}{$\mathcal{O}$}
\newcommand{\taskt}{$\mathcal{T}$}
\newcommand{\inferences}{$\mathcal{I}$}
\newcommand{\action}{$\mathcal{A}$}
% \aman{I agree with Jay that there is some disconnect. Maybe we can make things clearer by giving a simple example inline:}

% \aman{
In grounded social scenarios, an agent's perspective can be distilled into four primary variables:
\begin{enumerate*}
\item \textit{Observations} \obs~(e.g., \textit{Tom entered the kitchen. Tom wants a chocolate. Ella moves the chocolate.}),
\item \textit{Task} \taskt~(e.g., \textit{Based on the above observations, who needs help?}),
\item \textit{Inferences} \inferences~(e.g., \textit{Tom is unaware of the chocolate's current location.}), and
\item \textit{Action} \action~(e.g., \textit{Inform Tom about the chocolate's location.}).
\end{enumerate*} For a comprehensive illustration of these variables in context, please refer to Figure~\ref{fig:motivation}.
% }

% \amanx{Most existing social reasoning tasks evaluate models by asking questions about a specific inference step such as ``\textit{Where will Sally look for the marble?}'' with a list of candidate answers~\citep{nematzadeh2018evaluating, sap2019social, le2019revisiting}, as shown in Figure~\ref{fig:motivation} left side.
% Thus, the task can be expressed as $P(\mathcal{I}|\mathcal{O}, \mathcal{T}_I)$, where $\mathcal{T}_I$ refers to the inference-probing task verbalized as the specific question and inference candidates.
% We argue, however, that in real-life AI applications such as embodied agents, inferences are implicit and decisions are made in the realm of actions.
% We propose \textit{Thinking for Doing (T4D)} that evaluates model's capabilities to select the proper action given observations without instructing a specific inference direction.
% To make models reason about action, the task needs to shift from directly probing inference ($\mathcal{T}_I$) to asking about actions ($\mathcal{T}_A$).}{suggested rewrite}

% \aman{
Traditional social reasoning tasks typically challenge models with questions targeting specific inferences. 
For example, they might pose a question like ``\textit{Where will Jackson look for the onion?}'' accompanied by a set of candidate answers~\citep{nematzadeh2018evaluating, sap2019social, le2019revisiting}. 
This is depicted in the left side of Figure~\ref{fig:motivation}. 
Formally, this kind of task can be represented as estimation of $P(\mathcal{I}|\mathcal{O}, \mathcal{T}_I)$, where $\mathcal{T}_I$ denotes the inference-directed task articulated by the specific question and its associated answer options.
% }

% \aman{
However, in many real-world AI applications, particularly for embodied agents, decisions often revolve around actions rather than explicit inferences. 
These decisions are influenced by underlying, often \textit{implicit}, inferences. 
To bridge this gap, we introduce \textit{Thinking for Doing (T4D)}, a task designed to assess a model's ability to determine the appropriate action based solely on observations, without being directed towards a particular inference. 
Effectively, T4D represents a shift from directly probing for specific inferences ($\mathcal{T}_I$) to eliciting actions ($\mathcal{T}_A$).
% }
In the T4D framework, the model's task is not simply to make an inference but to decide on an action based on inferred mental states. 
This decision-making process involves estimating $P(\mathcal{A}|\mathcal{O}, \mathcal{T}_A)$, where $\mathcal{T}_A$ encapsulates the action-oriented task, such as determining \textit{Who would you prefer to assist the most?} with potential actions $\mathcal{A}$ like \textit{Assist Jackson} or \textit{Assist Noah}.
Crucially, in T4D, inferences $ \mathcal{I} $ act as a \textit{latent variable}, inferred from the observable $ \mathcal{O} $ to subsequently influence the chosen action $ \mathcal{A} $, \textit{i.e.} $P(\mathcal{A}|\mathcal{O}, \mathcal{T}_A, \mathcal{I})$.
\begin{wrapfigure}{r}{0.6\textwidth}
\vspace{-0.6cm}
  \begin{center}
    \includegraphics[width=0.6\textwidth]{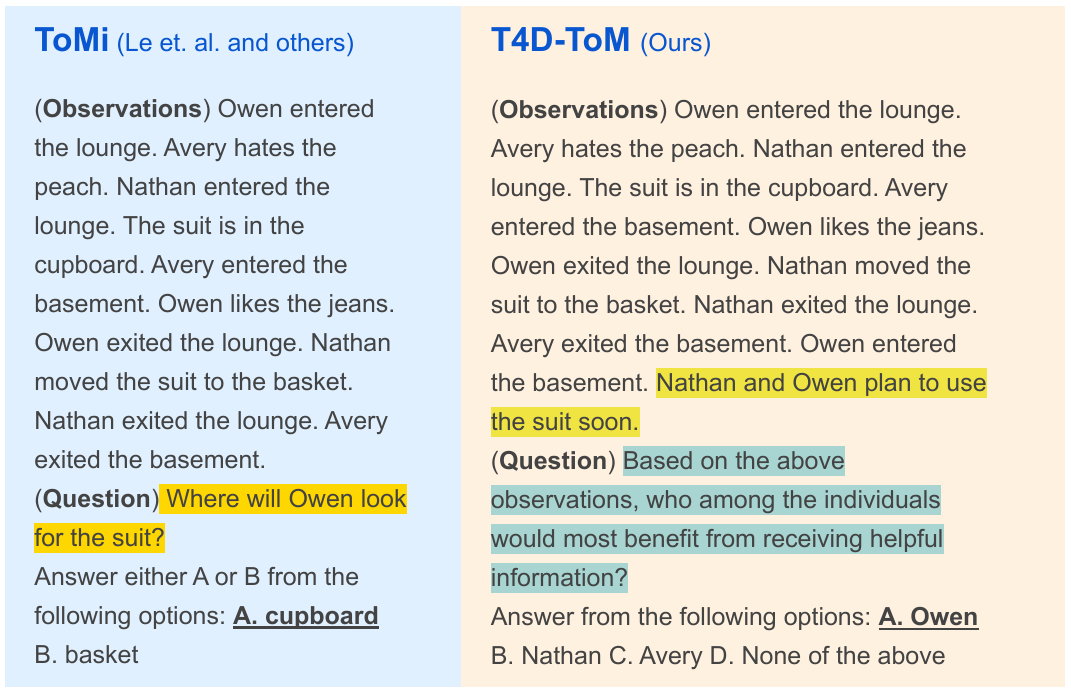}
  \end{center}
\vspace{-0.3cm}
  \caption{\small
	\textbf{Task input comparison} of ToMi that asks an inference question given observations and our converted \task that requires models to choose an action}
  \label{fig:data_example}
\vspace{-0.5cm}
\end{wrapfigure}

\subsection{Converting ToM Benchmarks to T4D}\label{sec:3.2}
This study focuses on a critical ability in social intelligence--Theory of Mind (ToM) and converts a widely-used existing benchmark: ToMi~\citep{le2019revisiting} from probing inferences to probing agent's action decisions.
% \paragraph{ToMi Dataset}
In the classic Sally-Anne Test setup (used by ToMi), participants interpret a stroy. 
For instance, consider Owen who mistakenly believes the suit is placed in the cupboard (Figure~\ref{fig:data_example}). 
ToMi asks models to deduce Owen's mental states, with the expected answer being that Owen will search for the suit inside the cupboard (due to mistaken beliefs).
% \aman{Clarify in the previous line: this is the ToMi task.}
% \paragraph{From ToMi to T4D}

To shift the focus towards actions as an agent who could potentially intervene and help other characters, we introduce an intent: \textit{both Owen and Nathan intend to use the suit in the near future}. 
% \amanx{We also mention that Nathan intends to use it to avoid models picking up the shortcut that they should help whoever plans to use something.}{
By explicitly stating both characters' intentions, we aim to deter models from adopting a rudimentary heuristic, like automatically assisting the character with immediate plans. 
However, we also ensure that this complexity does not obfuscate the task for humans.
% }
% \aman{Add: \textit{However, we also ensure that this complexity does not obfuscate the task for humans.}}
% \amanx{As we later show in human study~\ref{sec:3.3}, adding Nathan's intent still ensures that humans agree that Owen should be helped among options.}{
As validated in section~\ref{sec:3.3}, despite the shared intent to use the suit, human consensus consistently identifies Owen as the one needing help due to his misunderstanding about the suit's location.
% }
% \srividya {can add 1 or 2 lines of why and how this new intent might make the task better for LLMs}
In our modified task, termed \task, models are prompted to identify which character they would assist the most by providing accurate information about the onion's location.
% \amanx{To be able to choose the correct answer ``\textit{Owen}'' in T4D, the model has to understand that 1) Owen holds a false belief regarding the suit's location; and 2) given Owen's imminent need for the suit (emphasized in our setup), he stands to gain the most from the correct information.}{
Thus, in the T4D adaptation, models must deduce from the narrative that: 1) Owen remains under the false impression that the suit is in the cupboard, and 2) considering his impending need for the suit, accurate knowledge about its location would significantly benefit him.
%}
We programmatically convert the stories of ToMi (around 500) to \task due to ToMi's templatic nature.
Details of conversion are in Appendix~\ref{appdx:conver_detail}.

\subsection{Human Agreement on T4D}\label{sec:3.3}
Before using T4D to evaluate our models, we seek to verify its validity by testing it with human ToM (\textit{e.g.}, would human ToM encourage helping a character who holds outdated beliefs?).
% \amanx{We sample 76 instances (around 1/3 of the subset of ToMi that we convert to T4D) and ask \textit{20} human raters to answer the questions.}{
To do so, we randomly sampled around 80 instances for evaluation by $n = 20$ human raters.
To ensure this human study reflects how most people would use ToM in real life, we do \textit{not} pre-train these raters extensively on the ToM tasks and do \textit{not} provide any answers in the task examples. 
% Results showing human agreement are shown in Table~\ref{tab:agreement}.
% \amanx{We find that all instances have at least 17 out of 20 raters agree on the correct answer option. Specifically, around \textit{90\% of the instances reach more than 95\% agreement} (19 or more raters agree out of 20).}{
Our findings underscore the robustness of T4D tasks: every instance garnered agreement from at least 17 of the 20 raters. 
Moreover, \textit{over 90\% of the instances achieved agreement levels exceeding 95\%} (19 or all 20 raters in consensus).
% \amanx{Hence, we find that humans have high agreement on T4D tasks to determine a character to help.}{
This strong human consensus shows that the design of T4D naturally aligns with human perspectives on decision-making.

%% file: sections/4_T4D_perf.tex
% Human evaluation details and statistics
Here we test LLMs on our T4D task and compare with their performance on the original ToMi set that we convert from.
We use PaLM 2~\citep{anil2023palm} Bison (S) and Unicorn (L)~\footnote{\url{https://blog.google/technology/ai/google-palm-2-ai-large-language-model/}}, ChatGPT (GPT-3.5)~\citep{chatgpt}, and GPT-4~\citep{openai2023gpt} accessed between June and August, 2023.

\subsection{Thinking Is ``\textit{Easy}'', T4D Is Challenging for LLMs}
% \aman{How about calling this subsection: \textit{ToMi is easy, but T4D is Challenging}, as both the points are imp.}

\begin{wraptable}{r}{0.5\textwidth}
% \vspace{-0.7cm}
% \pei{REMINDER: Change to Center or Figure if Wraptable is not allowed!}
\caption{ \small \textbf{LLMs' accuracy on T4D compared with ToMi}. We find gap between human performance on T4D is much larger than that on ToMi (*we count humans correct when there is more than 95\% agreement).} 
\resizebox{0.5\textwidth}{!}{
\centering
\begin{tabular}{l|c|c}
\hline
\rowcolor[HTML]{E8EAED} 
\multicolumn{1}{c|}{\cellcolor[HTML]{E8EAED}{\color[HTML]{3C4043} \textbf{Models}}} & \multicolumn{1}{c|}{\cellcolor[HTML]{E8EAED}{\color[HTML]{3C4043} \textbf{ToMi}}} & \cellcolor[HTML]{E8EAED}{\color[HTML]{3C4043} \textbf{T4D-ToM}} \\ \hline
PaLM 2-S (Bison)                                                                    & \multicolumn{1}{c|}{87}                                                 & 16                                                        \\
PaLM 2-L (Unicorn)                                                                  & \multicolumn{1}{c|}{87 }                                                 & 30                                                         \\
GPT-3.5-turbo (ChatGPT)                                                             & \multicolumn{1}{c|}{74 }                                                     & 15                                                         \\
GPT-4                                                                               & \multicolumn{1}{c|}{\textbf{93}}                                             & \textbf{50}                                               \\
Random Guessing                                                                     & \multicolumn{1}{c|}{50 }                                                     & 26                                                         \\ \hline
\rowcolor[HTML]{E8EAED} 
\textbf{Human}                                                                      & \textbf{100}                                                                      & \textbf{90}*                                                             \\ \hline
\end{tabular}
}

\label{tab:sittom_hard}
\vspace{-0.3cm}
\end{wraptable}
We focus on zero-shot performance following recent studies~\citep{sap-etal-2022-neural, shapira2023clever,sclar2023minding} to probe LLM's capabilities to understand and use theory-of-mind.
Specifically, we provide answer options and instruct models to output one answer option.
% \aman{This evaluation approach requires models to make decisions based solely on their pre-training, without any task-specific fine-tuning or examples.}
The results comparing LLM's performance on ToMi and T4D-ToM are shown in Table~\ref{tab:sittom_hard}.
We find that both PaLM 2 and GPT models perform close to perfect human scores on ToMi (best model GPT-4 gets 93\% vs human 100\%) but the performance gap enlarges significantly across all models when tested on T4D-ToM (GPT-4 50\% vs human 90\%).
This discrepancy underscores the challenges posed by T4D for even the strongest contemporary LLMs.

% missing: structure in inference space
\subsection{What Makes T4D Challenging for LLMs?}\label{sec4.2:ablations}

% \amanx{To better understand \textit{why} LLMs find T4D challenging, we collect human-written rationales for solving the task, identify three dimensions of reasoning challenges, and make the task \textit{easier} for models by providing oracle reasoning steps corresponding to each of the three dimensions.}{
To better understand \textit{why} LLMs find T4D challenging, we conducted a study to understand the reasoning processes that humans use to tackle T4D tasks. By collecting and analyzing human-written rationales, we identified distinct dimensions of reasoning that seem particularly challenging for LLMs. Next, we discuss these challenges and experiments with oracle hints to determine if they can indeed aid the models in overcoming these reasoning hurdles.
% \amanx{From human rationales (examples in Appendix~\ref{appdx:human_study}), we find the following three major reasoning challenges with more details shown in Table~\ref{tab:reasoning_ablations}}{ 
The major reasoning challenges, along with examples and our proposed oracle hints, are summarized in Table~\ref{tab:reasoning_ablations} and we include example rationales in Appendix~\ref{appdx:human_study}.

% \begin{wraptable}{r}{0.7\textwidth}
% % \vspace{-0.6cm}
% \resizebox{0.7\textwidth}{!}{
\begin{table}[]
\caption{\textbf{Reasoning-Level breakdown}. Following the example task from Figure~\ref{fig:data_example}, we show 3 types of reasoning challenges with example specific reasoning steps and design oracle hints to make each challenge \textit{easier} to analyze what makes LLMs struggle on T4D.}
\centering
\resizebox{0.9\textwidth}{!}{
\begin{tabular}{c|l|l}
\hline
\textbf{\begin{tabular}[c]{@{}c@{}}Reasoning \\ Challenges\end{tabular}}  & \multicolumn{1}{c|}{\textbf{Example Reasoning Steps}}                                                                                                                                                                 & \multicolumn{1}{c}{\textbf{How to Provide Oracle Hints}}                                                                                                                         \\ \hline
\begin{tabular}[c]{@{}c@{}}Question \\ Decomposition (QD)\end{tabular}   & \textit{\begin{tabular}[c]{@{}l@{}}Who would benefit from info?\\ --\textgreater Nathan and Owen plan to use the suit\\ --\textgreater Do they know \textbf{the suit's location}?\end{tabular}}               & \begin{tabular}[c]{@{}l@{}}Add hint after question:\\ "HINT: this information is about\\  an \textbf{item’s location}"\end{tabular}           \\ \hline
\begin{tabular}[c]{@{}c@{}}Theory-of-Mind \\ (ToM)\end{tabular}          & \textit{\begin{tabular}[c]{@{}l@{}}Nathan and Owen plan to use the suit soon\\ --\textgreater They need to know the location\\ Owen left before the suit was moved\\ --\textgreater \textbf{Owen thinks the suit is in the cupboard}\end{tabular}} & \begin{tabular}[c]{@{}l@{}}Provide oracle ToM inference:\\ "\textbf{Owen will look for the suit in} \\ \textbf{the cupboard}"\end{tabular}                                 \\ \hline
\begin{tabular}[c]{@{}c@{}}Common Sense \\ Assumption (CSA)\end{tabular} & \begin{tabular}[c]{@{}l@{}}Nathan moved the suit to the basket\\ --\textgreater Though not mentioned, we can \\ \textbf{assume that the basket is lounge}\\ \textbf{as Nathan is not said to exit the room}\end{tabular}    & \begin{tabular}[c]{@{}l@{}}Make assumptions explicit:\\"\textbf{Cupboard and basket are in lounge}"\\"\textbf{Characters do not leave room}\\ \textbf{unless explicitly stated}"\end{tabular} \\ \hline
\end{tabular}
}

\label{tab:reasoning_ablations}
% \vspace{-0.5cm}
\end{table}
% \end{wraptable}

\textbf{Question Decomposition (QD)} We find that humans often break down the overarching T4D task into more specific follow-up questions such as ``\textit{Who might have an information gap?}'' and ``\textit{What information I can provide?}''.
% \amanx{that link the general question about action choice to potential relevant observations.}{
This decomposition bridges the gap between the general question and the provided observations.
% \amanx{We provide the QD reasoning steps by adding hints about the specific information the model can provide, specifically the \textit{oracle} inference results ($\mathcal{I}_Q$) from decomposing questions, \textit{i.e}, $P(A|\mathcal{O}, \mathcal{T}_A, \mathcal{I}_Q)$.}{
To emulate this in models, we added oracle hints, spotlighting specific information, derived from the decomposition process. Essentially, we guide the models with \textit{oracle} inference results ($\mathcal{I}_Q$), restructuring the task as \textit{i.e}, $P(A|\mathcal{O}, \mathcal{T}_A, \mathcal{I}_Q)$.

\textbf{Theory-of-Mind Inferences (ToM)} The second major reasoning challenge is the core inference tested in the Sally-Anne test -- can models correctly infer that Sally will look for the item in the \textit{old} location because she left the room before Anne moved the item? 
We make the \textbf{ToM} reasoning challenge easier by providing oracle ToM inferences ($\mathcal{I}_{ToM}$) in the observations: ``\textit{Sally will look for the [ITEM] in the [OLD CONTAINER]}''. 
This shifts the task to $P(A|\mathcal{O}, \mathcal{T}_A, \mathcal{I}_{ToM})$.

\textbf{Common Sense Assumptions (CSA)} The ambiguity inherent in ToMi, as noted by \citet{sclar2023minding}, presents another challenge.
To solve the task, models must assume that both containers are located in the room, even though this is never mentioned explicitly in the observations.
We make these assumptions explicit in the observations, \textit{i.e}, $P(A|\mathcal{O}, \mathcal{T}_A, \mathcal{K}_{CS})$, where we use $\mathcal{K}_{CS}$ to indicate commonsense knowledge not explicitly present in the observation.

\paragraph{Analysis Results}
\begin{wrapfigure}{r}{0.5\textwidth}
\vspace{-0.6cm}
  \begin{center}
    \includegraphics[width=1\linewidth]{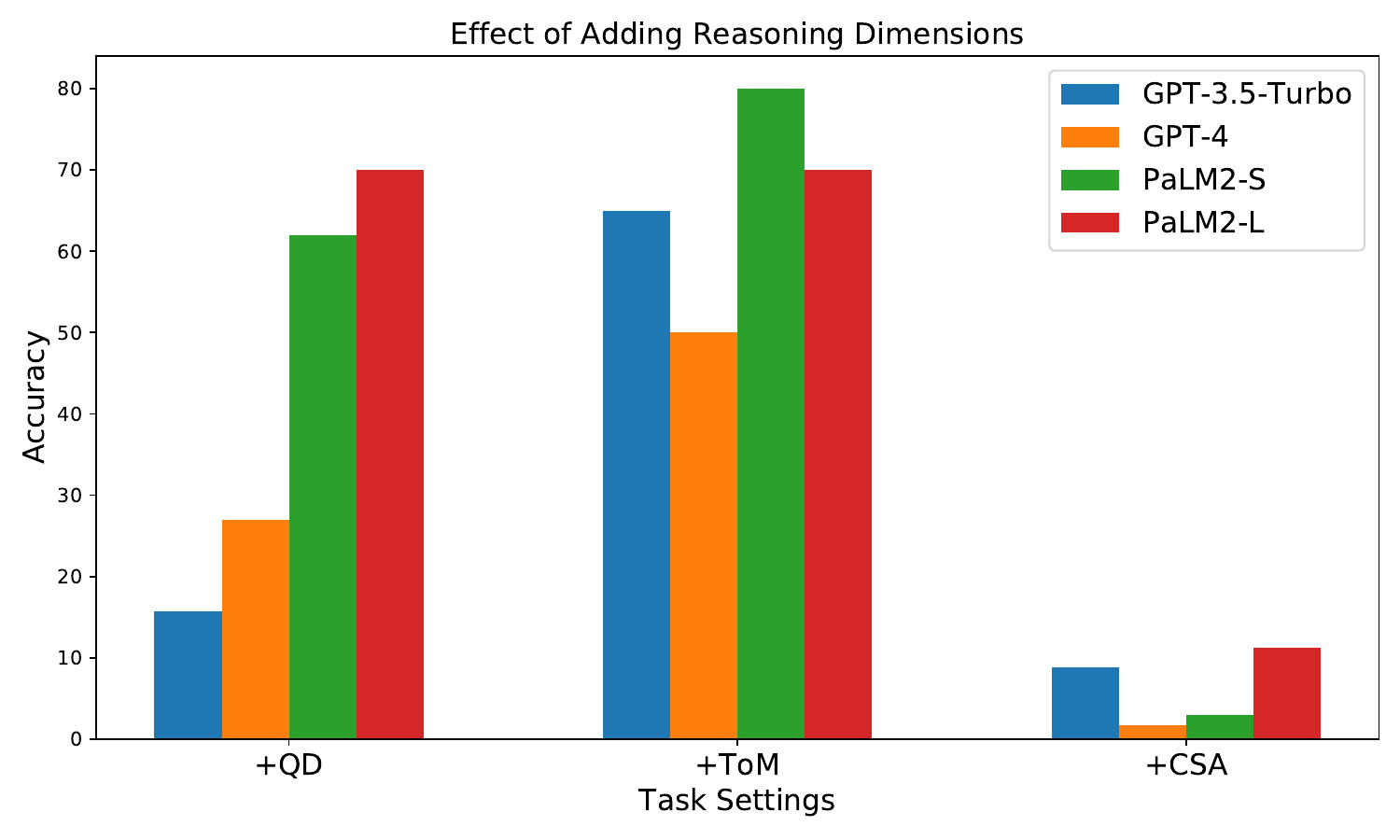}
  \end{center}
\vspace{-0.3cm}
	\caption{ \small \textbf{Increase in performance with provided reasoning levels}. Adding oracle inferences about question decomposition (especially for PaLM2) and ToM dramatically improve performance.}
	\label{fig:reasoning_levels}
 \vspace{-0.8cm}
\end{wrapfigure}
As illustrated in Figure~\ref{fig:reasoning_levels}, providing oracle hints yields varying results across the identified reasoning dimensions. Guiding models with hints related to item location (+QD) and incorporating oracle-derived character beliefs (+ToM) significantly enhances task performance. In contrast, merely clarifying assumptions (+CSA) has a limited effect on boosting model accuracy.

We hypothesize that providing QD or ToM inferences helps models by supplying \textit{suggestive evidence}, either in the form of leading questions ($\mathcal{I}_Q$) or relevant ToM inferences ($\mathcal{I}_{ToM}$).
These results also suggest that the underlying reason for the low performance of LLMs on T4D is attributed not to the task design but to their failure in drawing correct inferences and reasoning.
Thus, a key bottleneck in LLMs that makes T4D challenging (but easy for humans) is navigating the \textit{unconstrained} latent inference space $\mathcal{I}$ to \textit{locate the proper inference} that makes choosing which action intent clear.
% \amanx{Since if we provide hints on which is the relevant inference through ablation (providing $\mathcal{I}_Q$ or $\mathcal{I}_{ToM}$), models can arrive at the correct option close to human performance.}{}\aman{We have already made this point.}

% \aman{This is very good, and maybe should be referenced in the motivation more clearly: \textit{As we will show in Section ..., the struggle with T4D can be largely attributed to the failure of models to infer the right evidence.}}

%This also indicates that the reason why LLMs performance drop significantly in T4D is not solely because answering question about actions is unfamiliar to models' training corpora as we can see the increase with oracle inferences.

 % \pei{maybe adjust the fig to also show the final accuracy with oracle? sometimes improvement seems small because baseline is already high (GPT-4)}
 % The key bottleneck seems to be the combination of decomposing the question to connect with ToM reasoning (QD+ToM).
 % \aman{Won't it be clearer if we do + QD etc., as we are providing this information? Also, looks like ToM is adding some models way more?} \aman{Is there a \textbf{+everything} analysis?}

%% file: sections/5_FaR.tex
% \begin{figure}[h]
% 	% \centering
%     % \vspace{-0.2cm}
% 	\includegraphics[width=0.9\columnwidth]{figures/settings.pdf}
% % 	\vspace{-0.1cm}
% 	\caption{%\vspace{-0.5cm} 
%     \small
% 	Setting comparison of ToMi, T4D, ablations of T4D discussed in Section~\ref{sec4.2:ablations}, and our proposed FaR that gives LLMs heuristics to navigate latent inference space.}
% 	  	 % \vspace{-0.5cm}
% 	\label{fig:setting_comparison}
% \end{figure}

% Admissible of A*: heuristics chosen have this property to guranteered to 
% Heuristics search

% \pei{following text is one way to to frame our prompting method: guiding models to self-generate the path of connecting ToM thoughts and intents to help others, related to psychology traits.
% Alternatives (can be combined) include: connect to searching algorithms such as Astar and other formal reasoning}
% \aman{Pei, please repeat the argument from related work here.}

% \amanx{Here}{
Building on the insights from our T4D-ToM task analysis, we investigate can we help LLMs \textit{identify an implicit inference path} that leads to correct action choices without hints.
Given observations, humans find it natural to identify relevant inferences and arrive at decisions such as ``\textit{who should I provide information to?}'' 
However, ensuring that LLMs perform similarly structured reasoning is challenging. 
Although evidence points to LLMs' ability to infer, they do not necessarily connect these inferences to coherent reasoning about actions. 

\begin{figure*}[t!]
	\centering
    % \vspace{-0.4cm}
	\includegraphics[width=\linewidth]{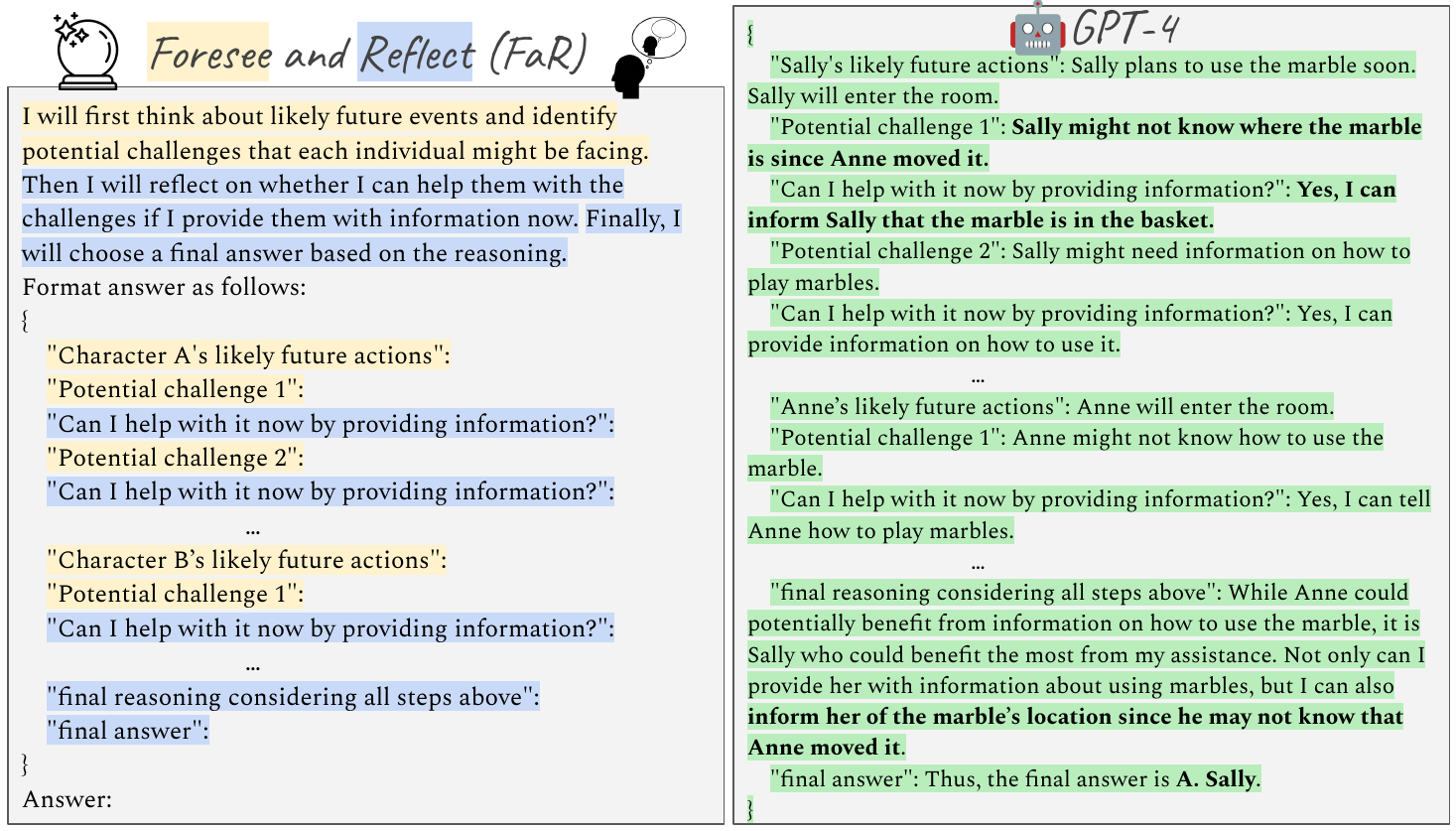}
% 	\vspace{-0.1cm}
	\caption{%\vspace{-0.5cm} 
    \small
	\textbf{\small \frameworklong~(\framework)} prompt (left), a new zero-shot prompting framework that combines \textit{future prediction} and pruning by \textit{action-aware reflection}. The \textit{Foresee} part is highlighted in yellow, \textit{Reflect} is highlighted in blue. Example GPT-4 output shown on the right. The model follows FaR and structures intermediate reasoning steps by copying keys and filling in the values so we only need \textit{one} inference call.}
	  	 \vspace{-0.7cm}
	\label{fig:FaR}
\end{figure*}

Our main methodology is to provide LLMs with a generalizable \textit{reasoning structure} that guides the models to relevant inferences.
% \amanx{We propose~\frameworklong~(\framework) framework to prompt LLMs to 1) consider potential future events based on partial observations and 2) reflect on what actions to perform at the moment to better help humans as an assisting agent.}{
To this end, we introduce the \textit{Foresee and Reflect} (\framework) framework. 
This framework equips LLMs with a structured reasoning paradigm, prompting them to: 1) extrapolate potential future events from given observations, and 2) introspect on actionable steps that would best serve humans in real-time contexts.
As argued in Section~\ref{rel_work}, the primary contribution of FaR is not to introduce a new prompt but to showcase the benefits of imposing a structured framework on the LLM’s reasoning process.
Figure~\ref{fig:FaR} presents FaR with an example output from GPT-4.

% \aman{IMO, we should completely delete the intuition subsection, and blend this with the first paragraph: \textit{Given observations, humans find it natural to identify relevant inferences and arrive at decisions such as ``\textit{who should I provide information to?}'' However, ensuring that LLMs perform similarly structured reasoning is challenging. Although evidence points to LLMs' ability to infer, they do not necessarily connect these inferences to coherent reasoning about actions. }}

% \subsection{\textit{Intuition}: Guiding Generative Models with Reasoning Structure to Extract Signals from Themselves}
% Given observations, humans find it natural to identify relevant inferences and arrive at \amanx{action}{} decisions such as ``\textit{who should I provide information to?}'' 
% % How can we teach generative models to do the same?
% \amanx{Evidence suggests LLMs are capable of inferences. Our experiments show that connecting inferences and reasoning does not happen automatically. We hypothesize a specific structure can enable LLMs to reason successfully across many, diverse scenarios.}{However, ensuring that LLMs perform similarly structured reasoning is challenging. Although evidence points to LLMs' ability to infer, they do not necessarily connect these inferences to coherent reasoning about actions.}
% % Our assumption is that LLMs can make such connections, but need to adjust the ``\textit{mindset}'' of how to reason when faced with choices of actions, since such scenarios rarely appear in training corpora, but are crucial for building a language agents.
% Thus, we could guide LLMs to explicitly generate reasoning steps connecting ``\textit{inferences}'' with ``\textit{actions}'' by injecting a reasoning \textit{structure}.
% Instead of letting LLMs to generate intermediate steps in an open-ended way such as ``\textit{reasoning step-by-step}'', can we give models high-level heuristics on how to reason?

% what are we trying to say: Humans can go from observations to actions.

\subsection{\textit{Foresee}: Considering Potential Future Events}
We design FaR by first prompting models to \textit{look into the future} by considering potential events that are likely to happen.
This stems from the understanding that the most valuable help often aligns with shaping a more desireable future outcome more desirable.
This is also related to a personality trait referred as ``\textit{Consideration of Future Consequences} (CFC)'' in psychology~\citep{strathman1994consideration}, which is the ability to predict future consequences to inform current action decisions.
% \amanx{Specifically, based on the observations $\mathcal{O}$, FaR prompts LLMs to go over each character appearing in the observations and predict their \textit{likely future actions}. Furthermore, models need to identify several \textit{potential challenges} that each character might be facing.}{
Given the observations $\mathcal{O}$, FaR guides LLMs to iterate over each character in the narrative, predicting their \textit{likely future actions} and pinpointing the \textit{potential challenges} they might encounter. This approach effectively broadens the initial observations, extrapolating inferences about potential future events.
% This process can be considered as \textit{expanding} the given observations to inferences about future.
% \amanx{We hypothesize that explicitly generating likely future events and challenges helps LLMs better connect the given observations with inferences \textit{relevant} to the action choices, which is providing help to others, and thus serve as heuristics while exploring the latent inference space.}{}

\subsection{\textit{Reflect}: Reasoning about Actions}
After \textit{foreseeing} likely future events, we prompt models to \textit{reflect} on whether performing actions at the moment could help with the potential challenges identified in the first step. 
% characters might be facing in the future.
This process can be considered as \textit{pruning} the generated potential future inferences based on the available action options.
% \amanx{This corresponds to ``\textit{inform current action decisions}'' and help LLMs to connect relevant inferences about future with the intent action choices, thus completing the full path of \textit{Observation--Inferences--Action}.}{
Overall, FaR helps LLMs connect relevant inferences about future with the intended action choices, completing a reasoning chain spanning \textit{Observation--Inferences--Action}.

% \aman{I think we are indexing on CFC more than we should. It is a bit distracting, because as a reader I am not clear about what we're doing and what we are directly taking from CFC. I would say we can just make a connection in passing, and add CFC in the related work.}

\paragraph{Connection to the A* Search Algorithm}

The FaR methodology is conceptually analogous to the A* search algorithm~\citep{hart1968formal}, an algorithm for finding the optimal path in a weighted graph. 
We draw the following connections:
\textbf{Start and Goal}: FaR begins with observations and aims to arrive at an optimal action decision. 
\textbf{Expanding Nodes}: In the \textit{Foresee} phase of FaR, potential inferences (akin to nodes in A*) are expanded by considering future events.
\textbf{Heuristics}: The predictions made during the \textit{Foresee} step act as heuristics, guiding the reasoning process toward the most relevant inferences.
\textbf{Path Pruning}: The \textit{Reflect} stage in FaR narrows down the inferred events based on available actions, similar to how A* prunes paths based on the heuristic and cost so far.

% \begin{itemize}
%     \item \textbf{Start and Goal}: FaR begins with observations and aims to arrive at an optimal action decision.
%     \item \textbf{Expanding Nodes}: In the \textit{Foresee} phase of FaR, potential inferences (akin to nodes in A*) are expanded by considering future events.
%     \item \textbf{Heuristics}: The predictions made during the \textit{Foresee} step act as heuristics, guiding the reasoning process toward the most relevant inferences.
%     \item \textbf{Path Pruning}: The \textit{Reflect} stage in FaR narrows down the inferred events based on available actions, similar to how A* prunes paths based on the heuristic and cost so far.
% \end{itemize}

% \begin{algorithm}
% \caption{FaR Reasoning Inspired by A*}
% \label{alg:far_a_star}
% \begin{algorithmic}[1]
% \STATE \textbf{Input:} Observations \( \mathcal{O} \), Action choices \( \mathcal{A} \)
% \STATE Initialize open list \( \textit{OPEN} \) with start node (initial observations)
% \STATE Initialize closed list \( \textit{CLOSED} \) as empty

% \end{algorithmic}
% \end{algorithm}

%% file: sections/6_experiments.tex
We examine the potential of various zero-shot prompting methods on improving LLM's performance on T4D and conduct generalization tests.
We aim to answer three research questions through our experiments: 1) \textit{How much can FaR improve LLM's zero-shot performance on T4D?} 2) \textit{Are both the ``foresee'' and ``reflect'' components necessary, and what are the limitations of FaR?} and 3) \textit{Does FaR generalize robustly across scenarios where models need to connect inferences with intents?}

\begin{figure}[t!]
	\centering
	\vspace{-0.3cm}
 \includegraphics[width=0.9\columnwidth]{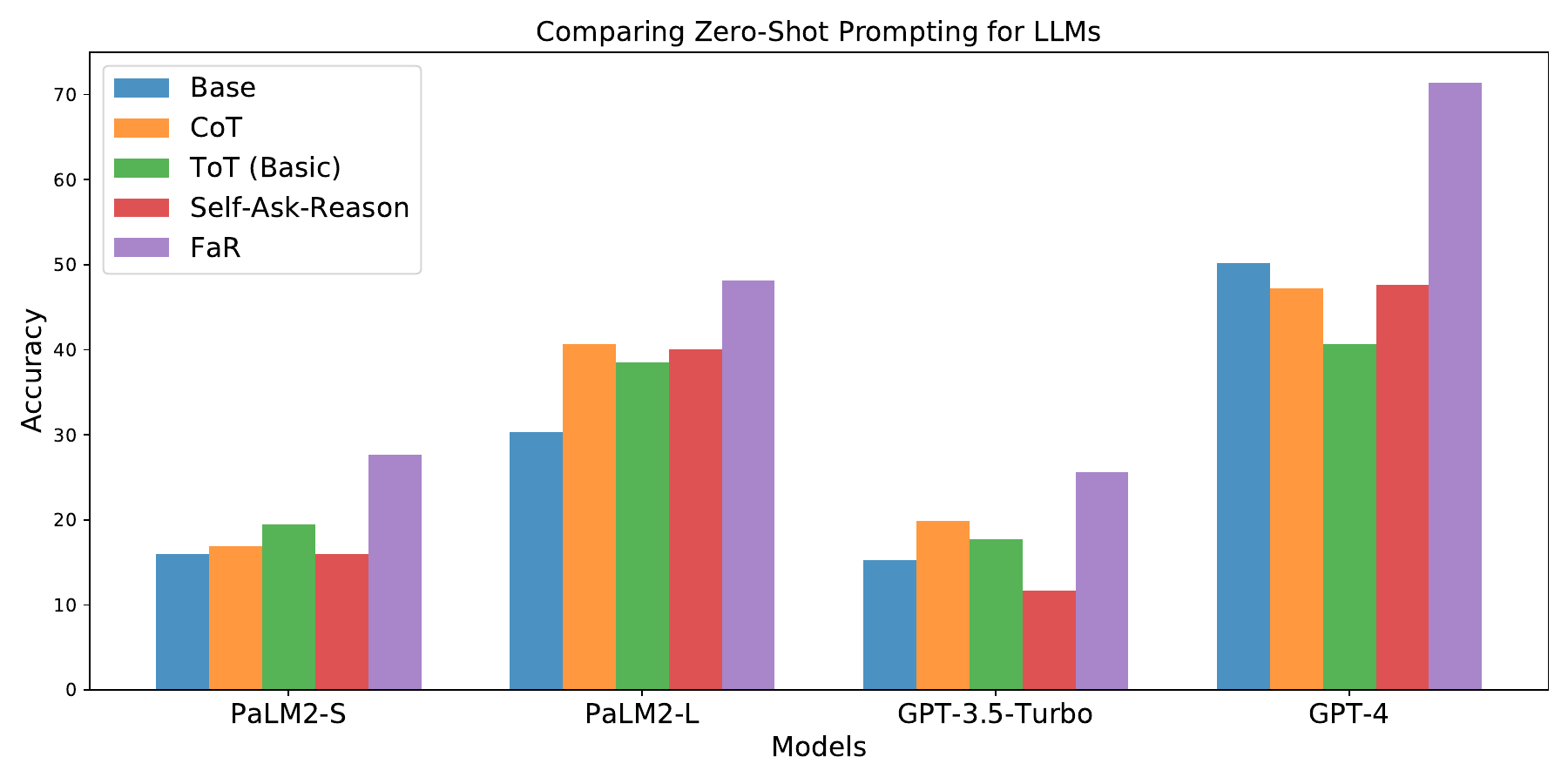}
	\vspace{-0.3cm}
	\caption{ \textbf{Comparison of zero-shot prompts.} We find FaR improves LLMs performance the most.}

	\label{fig:FaR_LLMs}
 \vspace{-0.5cm}
\end{figure}

\subsection{Baselines}\label{sec:6.1}
We consider the following zero-shot prompting strategies, each offering a unique reasoning structure. Full descriptions of the prompts are available in the Appendix~\ref{appdx:prompts}
% We consider the following zero-shot prompts with example full prompts shown in Appendix~\ref{appdx:prompts}:
\textbf{Chain-of-Thought (CoT)}~\citep{wei2022chain}:the zero-shot variant from~\citet{kojima2022large} and add ``\textit{Answer this question by reasoning step-by-step.}''
\textbf{Tree-of-Thought (ToT)}~\citep{yao2023tree} (Basic Zero-Shot): a zero-shot variant inspired by ToT, which prompts the LLM to envision a discussion among experts. Each expert contributes a reasoning step, and if any expert detects an error in their logic, they exit the discussion. 
% are discussing the problem and at each step, an expert shares a step of their thinking, then all experts go with the next steps, and if any expert realizes that they are wrong, they leave the discussion.
\textbf{Self-Ask}~\citep{press2022measuring}: this method emphasizes self-inquiry. Models generate and answer their follow-up questions, iterating until they reach a conclusive answer. A final reasoning step solidifies the conclusion.
% we prompt models to ask follow-up questions and answer them and iterate the process until it gets a final answer. We also add a final reasoning step after the follow-up questions and answers and before the final answer.
\textbf{FaR}: following Section~\ref{FaR} and Figure~\ref{fig:FaR}, we design a prompt that guides models to think about likely future events and challenges that characters might encounter, and reflect whether they can provide help.
We apply each prompt and make \textit{one} inference call on all LLMs with maximum 800 tokens with a temperature of 0 (greedy sampling).

\subsection{\framework Dramatically Improves GPT-4 Zero-Shot Performance}\label{sec:6.2}
Figure~\ref{fig:FaR_LLMs} present results of 4 different zero-shot prompting methods.
We find that FaR can significantly boost LLMs' performance on T4D-ToM while other prompting methods do not help much.
Specifically, FaR helps increase GPT-4 accuracy from base 50\% to 71\% as well as all other LLMs with the improvement between 12\% and 18\%.
We also observe that more powerful models (GPT-4 and PaLM2-L) tend to benefit more from FaR.

\subsection{Ablation and Analysis}\label{sec:6.3}
\paragraph{Both Foresight and Reflection Are Important}
FaR consists of two main components, one to \textit{foresee} future events and challenges and one to \textit{reflect} on action decisions.
To investigate the individual impact of these components, we modified the FaR prompt, isolating each element for ablation. Specifically, we omitted the foresight (referenced as yellow text in Figure~\ref{fig:FaR}) and reflection parts (blue text in Figure~\ref{fig:FaR}). 
Table~\ref{tab:FaR_ablation} presents ablation on FaR for the two components using GPT-4. 
We find that the performance significantly drops 17 and 12 points, respectively, when there is no \textit{foresee} and there is no \textit{reflect}, indicating that they are both crucial for T4D.

\begin{wraptable}{r}{0.4\textwidth}
\vspace{-0.3cm}
\caption{FaR ablations.}
\resizebox{0.4\textwidth}{!}{

\begin{tabular}{l|c}
\hline
\rowcolor[HTML]{EFEFEF} 
\multicolumn{1}{c|}{\cellcolor[HTML]{EFEFEF}\textbf{Prompts}} & \textbf{GPT-4 Accuracy} \\ \hline
Base                                                                       & 50.2                      \\
FaR-\textbf{NoForesee}                                                           & 53.2                       \\
FaR-\textbf{NoReflect}                                                              & 59.7  
\\
FaR-\textbf{NoisyForesee}                                                              & 42 \\
FaR                                                                        & \textbf{71.4}              \\
Random Guessing                                                            & 26              \\
\hline
\rowcolor[HTML]{EFEFEF} 
\textbf{Human}                                                             & 90                        \\ \hline
\end{tabular}
}
\vspace{-0.5cm}
\label{tab:FaR_ablation}
\end{wraptable}

\paragraph{Providing Noisy Foresight Undermines Performance}
We further assessed the robustness of the FaR framework by introducing \textit{noisy} foresight.
% , where irrelevant or misleading inferences about future events were provided as part of the prompt. 
For instance, a spurious foresight for the example in Figure~\ref{fig:FaR} might be``\textit{Sally will enter the bedroom to sleep.}'' without any evident reason from the observations. 
Table~\ref{tab:FaR_ablation} shows that LLMs are very sensitive to manually-inputted reasoning steps in FaR and the accuracy of GPT-4 drops from 71\% to 42\% (even lower than baseline).
This highlights a limitation:   while the FaR framework can enhance reasoning when guided correctly, it's sensitive to the quality of the foresight provided and can degrade performance if misled.

% a limitation for FaR as a zero-shot prompt is that it is very challenging for LLMs to \textit{recover} from a noisy or irrelevant foresight.

\begin{wraptable}{r}{0.4\textwidth}
\vspace{-0.7cm}
\caption{ Results on story-structure tests. FaR consistently improves the most.}
\resizebox{0.4\textwidth}{!}{
\begin{tabular}{lcccc}
% \centering
\rowcolor[HTML]{CBCEFB}
\textbf{D1}                      & \multicolumn{1}{l}{\cellcolor[HTML]{CBCEFB}} & \multicolumn{1}{l}{\cellcolor[HTML]{CBCEFB}} & \multicolumn{1}{l}{\cellcolor[HTML]{CBCEFB}}                                                   & \multicolumn{1}{l}{\cellcolor[HTML]{CBCEFB}} \\
\rowcolor[HTML]{E8EAED} 
{\color[HTML]{3C4043} \textbf{Model}}          & {\color[HTML]{3C4043} \textbf{CoT}}          & {\color[HTML]{3C4043} \textbf{ToT}}    & {\color[HTML]{3C4043} \textbf{Self-Ask}} & {\color[HTML]{3C4043} \textbf{FaR}}          \\
\rowcolor[HTML]{FFFFFF} 
{\color[HTML]{5F6368} GPT-3.5} & {\color[HTML]{5F6368} \textbf{52}}        & {\color[HTML]{5F6368} 39}                 & {\color[HTML]{5F6368} 26}                                                                   & {\color[HTML]{5F6368} \textbf{52}}        \\
\rowcolor[HTML]{F1F3F4} 
{\color[HTML]{5F6368} GPT-4}                   & {\color[HTML]{5F6368} \textbf{71}}                 & {\color[HTML]{5F6368} 29}                 & {\color[HTML]{5F6368} 33}                                                                   & {\color[HTML]{5F6368} 56}                 \\
\rowcolor[HTML]{FFFFFF} 
{\color[HTML]{5F6368} PaLM 2-S}                & {\color[HTML]{5F6368} 69}                 & {\color[HTML]{5F6368} 85}                 & {\color[HTML]{5F6368} 52}                                                                   & {\color[HTML]{5F6368} \textbf{87}}        \\
\rowcolor[HTML]{F1F3F4} 
{\color[HTML]{5F6368} PaLM 2-L}                & {\color[HTML]{5F6368} 84}                 & {\color[HTML]{5F6368} 92}                 & {\color[HTML]{5F6368} 87}                                                                   & {\color[HTML]{5F6368} \textbf{92}}        \\
\rowcolor[HTML]{CBCEFB} 
\textbf{D2}                     & \multicolumn{1}{l}{\cellcolor[HTML]{CBCEFB}} & \multicolumn{1}{l}{\cellcolor[HTML]{CBCEFB}} & \multicolumn{1}{l}{\cellcolor[HTML]{CBCEFB}}                                                   & \multicolumn{1}{l}{\cellcolor[HTML]{CBCEFB}} \\
\rowcolor[HTML]{E8EAED} 
{\color[HTML]{3C4043} \textbf{Model}}          & {\color[HTML]{3C4043} \textbf{CoT}}          & {\color[HTML]{3C4043} \textbf{ToT}}    & {\color[HTML]{3C4043} \textbf{Self-Ask}} & {\color[HTML]{3C4043} \textbf{FaR}}          \\
\rowcolor[HTML]{FFFFFF} 
{\color[HTML]{5F6368} GPT-3.5} & {\color[HTML]{5F6368} 21}                 & {\color[HTML]{5F6368} 36}                 & {\color[HTML]{5F6368} 44}                                                                   & {\color[HTML]{5F6368} \textbf{70}}        \\
\rowcolor[HTML]{F1F3F4} 
{\color[HTML]{5F6368} GPT-4}                   & {\color[HTML]{5F6368} 36}                 & {\color[HTML]{5F6368} 34}                 & {\color[HTML]{5F6368} 60}                                                                   & {\color[HTML]{5F6368} \textbf{95}}        \\
\rowcolor[HTML]{FFFFFF} 
{\color[HTML]{5F6368} PaLM 2-S}                & {\color[HTML]{5F6368} 36}                 & {\color[HTML]{5F6368} 39}                 & {\color[HTML]{5F6368} 15}                                                                   & {\color[HTML]{5F6368} \textbf{42}}        \\
\rowcolor[HTML]{F1F3F4} 
{\color[HTML]{5F6368} PaLM 2-L}                & {\color[HTML]{5F6368} 27}                 & {\color[HTML]{5F6368} 15}                 & {\color[HTML]{5F6368} 22}                                                                   & {\color[HTML]{5F6368} \textbf{90}}        \\
\rowcolor[HTML]{CBCEFB} 
\textbf{D3}               & \multicolumn{1}{l}{\cellcolor[HTML]{CBCEFB}} & \multicolumn{1}{l}{\cellcolor[HTML]{CBCEFB}} & \multicolumn{1}{l}{\cellcolor[HTML]{CBCEFB}}                                                   & \multicolumn{1}{l}{\cellcolor[HTML]{CBCEFB}} \\
\rowcolor[HTML]{E8EAED} 
{\color[HTML]{3C4043} \textbf{Model}}          & {\color[HTML]{3C4043} \textbf{CoT}}          & {\color[HTML]{3C4043} \textbf{ToT}}    & {\color[HTML]{3C4043} \textbf{Self-Ask}} & {\color[HTML]{3C4043} \textbf{FaR}}          \\
\rowcolor[HTML]{FFFFFF} 
{\color[HTML]{5F6368} GPT-3.5} & {\color[HTML]{5F6368} 35}                 & {\color[HTML]{5F6368} 48}                 & {\color[HTML]{5F6368} 9}                                                                   & {\color[HTML]{5F6368} \textbf{50}}        \\
\rowcolor[HTML]{F1F3F4} 
{\color[HTML]{5F6368} GPT-4}                   & {\color[HTML]{5F6368} 79}                 & {\color[HTML]{5F6368} 76}                 & {\color[HTML]{5F6368} 63}                                                                   & {\color[HTML]{5F6368} \textbf{100}}        \\
\rowcolor[HTML]{FFFFFF} 
{\color[HTML]{5F6368} PaLM 2-S}                & {\color[HTML]{5F6368} 12}                 & {\color[HTML]{5F6368} 20}                 & {\color[HTML]{5F6368} 20}                                                                   & {\color[HTML]{5F6368} \textbf{73}}        \\
\rowcolor[HTML]{F1F3F4} 
{\color[HTML]{5F6368} PaLM 2-L}                & {\color[HTML]{5F6368} 46}                 & {\color[HTML]{5F6368} 37}                 & {\color[HTML]{5F6368} 12}                                                                   & {\color[HTML]{5F6368} \textbf{82}}       
\end{tabular}
}

\vspace{-0.7cm}
\label{tab:structure_change}
\end{wraptable}

\subsection{\framework Generalizes to Diverse Scenarios}\label{sec:6.4}
We probe the generalizability of \framework by evaluating its efficacy on \textit{out-of-distribution} scenarios.
% These include variations on story structure and non-False-Belief Test scenarios that require models to use ToM to choose whom to provide emotional support.

\paragraph{Story Structure Robustness Tests}
\begin{wraptable}{r}{0.23\textwidth}
\vspace{-0.6cm}
\caption{Faux Pas results using GPT-4.}
\resizebox{0.23\textwidth}{!}{
\begin{tabular}{l|c}
\hline
\rowcolor[HTML]{E8EAED} 
\multicolumn{1}{c|}{\cellcolor[HTML]{E8EAED}{\color[HTML]{3C4043} \textbf{Prompts}}} & {\color[HTML]{3C4043} \textbf{Accuracy}} \\ \hline
Base                                                                                 & 31\%                            \\
CoT                                                                                  & 39\%                           \\
ToT                                                                          & 36\%                                     \\
Self-Ask                                                                             & 43\%                                     \\
Few-Shot                                                                             & 41\%                                     \\
FaR                                                                                  & \textbf{76\%}                            \\ \hline
\end{tabular}
}
\vspace{-0.5cm}
\label{tab:faux_pas}
\end{wraptable}
% We test whether FaR can generalize to story structures other than those included ToMi and use the 
We use three challenge sets from~\citet{sclar2023minding} to test if FaR can generalize to story structures beyond those included ToMi.
% We test whether FaR can generalize to story structures other than those included ToMi and use the three challenge sets from~\citet{sclar2023minding}.
% where they add complexity in the story structure by including 2 items being moved in 2 rooms (D1), more characters move the item (D2), and the same item moved among 4 containers (D3)
These sets introduce complexities such as the relocation of two items across two rooms (D1), the involvement of multiple characters with an item (D2), and a single item's movement among four containers (D3)~\footnote{\citet{sclar2023minding} propose \textit{SymbolicToM}, a symbolic tracker of mental states for ToMi. We do not include SymbolicToM for comparison in T4D because including answers from the tracker gives away that the model should focus on inferences about \textit{item's location}, whereas other methods are not provided with such assumptions.}.
We convert each set (100 stories each) to T4D-style probes using our ToMi conversion methodology.
Table~\ref{tab:structure_change} shows results on three types of story-structure change of the ToMi stories.
Overall, FaR helps LLMs achieve the highest accuracy compared to other zero-shot prompts on all three generalization tests, for almost all models.
% This demonstrates that the benefits of FaR on LLMs can generalize to diverse story structures.

\paragraph{T4D-Faux Pas Case Studies}
% We examine whether \framework can help LLMs decide actions in more diverse social scenarios by testing on \textit{non-Sally-Anne-Test-like} context, \textit{i.e.}, when the proper action is \textit{not} about providing information to correct a false belief.
To further ascertain \framework's adaptability, we ventured beyond the classic Sally-Anne Test context. We explored Faux Pas scenarios, characterized by individuals inadvertently sharing potentially distressing or unwanted information~\citep{baron1999recognition}. 
We consider Faux Pas, a category of social stories where a person ``\textit{says something without considering if it is something that others might not want to hear or know}''~\citep{baron1999recognition}, and use 20 expert-curated stories from~\citet{shapira-etal-2023-well}.
We convert the original set to T4D by asking models to choose a character from the stories to provide \textit{emotional support} (examples Appendix~\ref{appdx:generalization}). 
We test GPT-4 with multiple zero-shot prompts as well as few-shot prompting with examples from T4D converted from ToMi.
Table~\ref{tab:faux_pas} shows that FaR outperforms other methods dramatically, showing the generalizability of the zero-shot prompt FaR.

%% file: sections/7_conclusion.tex
% We propose T4D to challenge LLMs to connect ToM reasoning to actions.
We propose T4D, a task designed to challenge the capacity of LLMs in bridging Theory of Mind reasoning to actions.
Our analyses highlighted a key limitation in LLMs: their difficulty in grappling with implicit inferences without explicit guidance.
To mitigate this, we introduced FaR, a structured reasoning paradigm, which not only boosts the performance of LLMs but also ensures broader generalization.
As a next step, it would be valuable to delve deeper into understanding the internal representation of LLMs when guided by structured prompts like FaR.
% Through analysis, we find that LLMs struggle to identify implicit inferences when not instructed and we propose FaR to inject a reasoning structure and find success in performance and generalization.

%% file: sections/9_appendix.tex
% \section{Frequently Asked Questions (FAQ)}\label{appdx:faq}
% \subsection{}

\section{ToMi Conversion Details}\label{appdx:conver_detail}
ToMi~\citep{le2019revisiting} was proposed as a question answering task based on Sally-Anne Tests and improved upon previous benchmark from~\citet{nematzadeh2018evaluating} by removing statistical biases making the task solvable without ToM.
Specifically, ToMi defines multiple story primitives such as ``\textit{A enters the room}'', ``\textit{B moves the item}'',``\textit{A left the room}'', etc. and primitives are combined into stories with a finite set of orderings~\citep{sclar2023minding}.
Prior work such as~\citet{sap-etal-2022-neural} has found some errors in the ToMi dataset and filtered a clean version that we use to convert to T4D.

On a high-level, conversion consists of two main changes: 1) we add a sentence at the end of the story with the intents of the two characters involved in moving the item (``\textit{Sally and Anne plan to use the marble soon.}''); 2) we propose a new question given the stories about a situated agent's action and provide a list of answer options from all the characters and a ``\textit{None of the Above}'' option. 
Specifically, we need to parse the original ToMi tasks to find out 3 types of characters to programmatically generate the additional sentence and new QA options: 1) the character who holds a false belief since they left before another character moves the item. This is also the character who is the correct answer of T4D task, as they benefit the most from receiving helpful information; 2) the character who moves the item after the first character left; 3) distracting characters who do not involve in moving or needing to use the item, as they were introduced in ToMi to reduce biases.

We extract these character names from raw ToMi stories by extracting entities before verbs such as ``\textit{moves}'', ``\textit{enters}'', ``\textit{leaves}'', etc.
Then we find the character holding a false belif by extracting from the \textit{original} ToMi questions such as ``\textit{Where will Sally look for the marble?}'', since the ToMi questions directly targets the character holding a false belief.
Next we find who is the character moving the item by extracting the name before ``\textit{moves}'', and the final character (if any) would be the distracting character.
Due to the templatic nature of ToMi, we extract these character names from all stories automatically.
Finally, with extracted character names of each type, we apply the changes mentioned above by filling the names.

\section{Human Study Details}\label{appdx:human_study}

\paragraph{T4D Task Setup}

\begin{figure}[t!]
	\centering
	% \vspace{-0.3cm}
 \includegraphics[width=0.8\columnwidth]{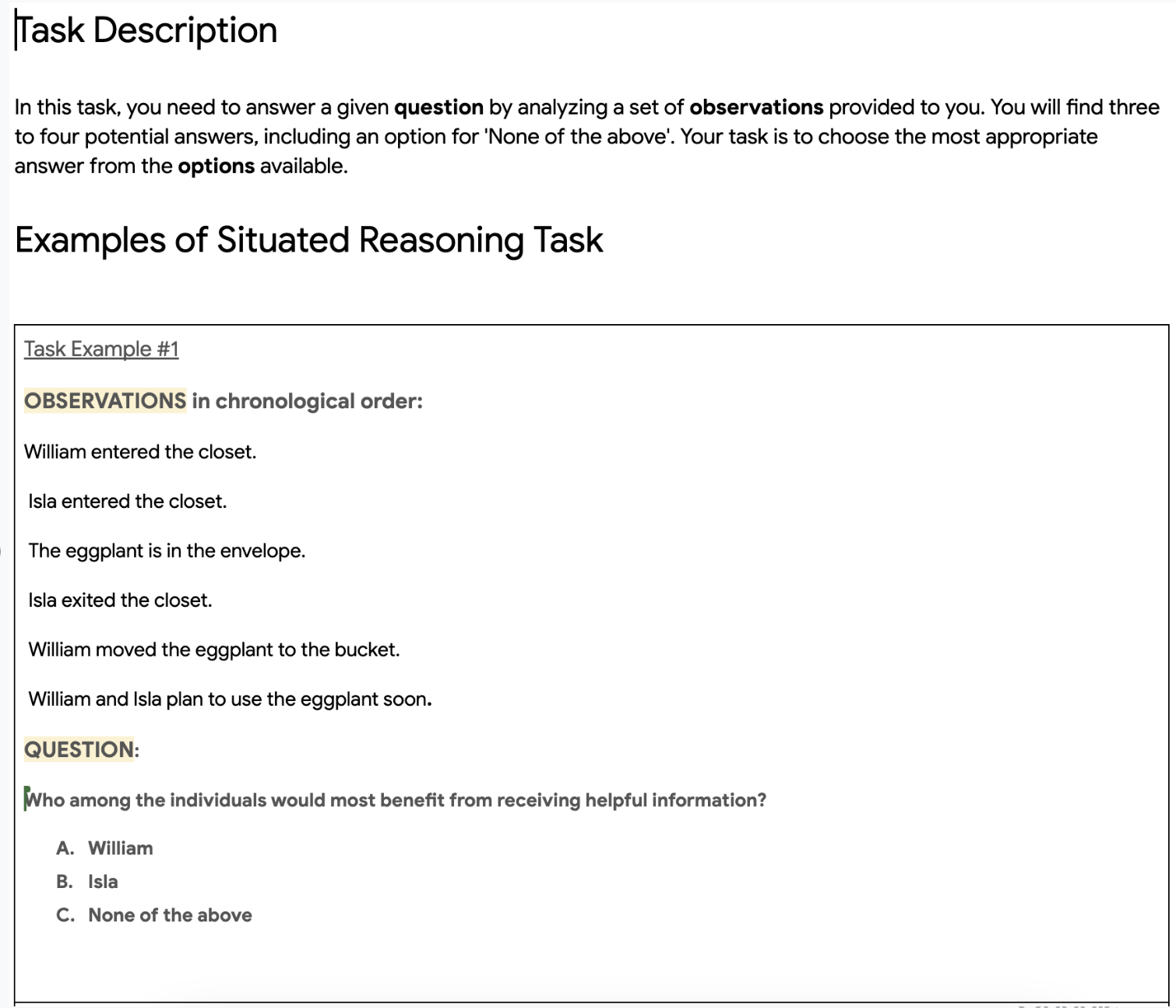}
	% \vspace{-0.3cm}
	\caption{ \small Human study instruction screenshot with example.}

	\label{fig:human_UI}
 % \vspace{-0.5cm}
\end{figure}

As mentioned in Section~\ref{sec:3.3}, we conduct human studies with 20 raters who are not trained on T4D tasks and not provided answers in examples.
The annotators were selected randomly from a large pool of in-house human annotators.
Figure~\ref{fig:human_UI} shows a screenshot of our instructions for human study with an example.
We provide 3 of such examples without answers and raters answer sampled T4D instances in the multi-choice QA format as well.
We also leave it optional for raters to provide their rationales of solving T4D.
% Human study consensus details shown in Table~\ref{tab:agreement}.

% \begin{wraptable}{r}{0.33\textwidth}
% % \vspace{-0.6cm}
% \resizebox{0.33\textwidth}{!}{
% \begin{tabular}{c|c|c}
% \hline
% \textbf{\begin{tabular}[c]{@{}c@{}}Consensus on\\ Correct Answer\end{tabular}} & \textbf{\#count} & \textbf{\%} \\ \hline
% 20/20                                                                          & 16               & 21.10\%     \\
% 19/20                                                                          & 52               & 68.40\%     \\
% 18/20                                                                          & 6                & 7.90\%      \\
% 17/20                                                                          & 2                & 2.60\%      \\ \hline
% \end{tabular}
% }
% % \vspace{-0.3cm}
% \caption{\small \textbf{Human Agreement} on 76 samples with 20 repetition ratings. Each row is one consensus distribution from the study. Over 89\% of instances have 95\% agreement.}
% % \aman{What are the rows? Option number?}}
% \label{tab:agreement}
% % \vspace{-0.2cm}
% \end{wraptable}

\paragraph{Examples of Human Rationales}
\begin{table}[b]
\centering
\caption{\small Examples of human rationales we collected from human study. We highlighted parts from the rationales to correspond to the 3 reasoning levels discussed in~\ref{sec4.2:ablations}: \textcolor{red}{question decomposition}, \textcolor{blue}{theory-of-mind inferences} (about others' goals and beliefs), and \textcolor{orange}{commonsense assumptions}.}
\resizebox{0.9\textwidth}{!}{
\begin{tabular}{c|l}
\hline
Rationale 1 & \begin{tabular}[c]{@{}l@{}}\textcolor{red}{Who in the stories have goals that require information?} We know that William and Isla \\ both need to use eggplant. \textcolor{blue}{They need to know the location of it to be able to use it.} \\ William moved the eggplant after Isla exited the closet, \textcolor{blue}{thus Isla is not aware of the }\\\textcolor{blue}{ current location.} I should choose Isla.\end{tabular} \\ \hline

Rationale 2 & \begin{tabular}[c]{@{}l@{}}William and Isla both plan to use the eggplant. But it is \textcolor{blue}{Isla who lacks the knowledge}\\ \textcolor{blue}{of the current location of the eggplant because William moved it} \textcolor{orange}{(assuming that both}\\ \textcolor{orange}{the envelop and the bucket are in the closet}. Thus the answer should be Isla.\end{tabular}                                                      \\ \hline

Rationale 3 & \begin{tabular}[c]{@{}l@{}}Isla is the right answer because she plans to use the eggplant \textcolor{blue}{ but she does not know} \\ that it has been moved to another location. \textcolor{red}{She benefits from me telling her to avoid}\\ \textcolor{red}{inconvenience}.\end{tabular}                                                                                                                      \\ \hline
\end{tabular}
}

\label{tab:rationale_examples}

\end{table}
In Section~\ref{sec4.2:ablations}, we summarize 3 reasoning levels from collected human rationales.
Table~\ref{tab:rationale_examples} presents examples with highlighted texts corresponding to each of the 3 reasoning levels.

\section{Prompt Details}\label{appdx:prompts}
Here we show different examples with full zero-shot prompts we use in our experiments.
As discussed, we make 1 inference/API call with each zero-shot prompt with maximum tokens of 800.
\paragraph{Chain-of-Thought}

\texttt{You are a clever and helpful AI assistant. Your goal is to help human users around you.}

\texttt{One day you have the following observations that follow chronological order: }

\texttt{Chloe loves the undershirt. Avery entered the dining room. Aiden entered the dining room. The stockings is in the crate. Avery exited the dining room. Aiden moved the stockings to the cupboard. Aiden exited the dining room. Avery entered the sunroom. Aiden and Avery plan to use the stockings soon.}

\texttt{\textbf{Answer the following multiple-choice question by reasoning step-by-step.}}

\texttt{Based on the above information, who among the individuals would most benefit from receiving helpful information? Answer from the following options: A. Chloe B. Avery C. Aiden D. None of the above.}

\texttt{End your answer with 'Thus, the final answer is A (or B or C or D)'}

\texttt{Answer:
}

\paragraph{Tree-of-Thought (Basic Zero-Shot)}

\texttt{You are a clever and helpful AI assistant. Your goal is to help human users around you.}

\texttt{One day you have the following observations that follow chronological order: }

\texttt{Sophia entered the front yard. Jayden entered the bathroom. Logan entered the front yard. The cabbage is in the suitcase. Jayden hates the spinach. Jayden exited the bathroom. Logan exited the front yard. Jayden likes the grapefruit. Sophia moved the cabbage to the basket. Sophia exited the front yard. Logan entered the bathroom. Sophia and Logan plan to use the cabbage soon.}

\texttt{\textbf{Imagine three different experts are answering this question.}}

\texttt{\textbf{All experts will write down 1 step of their thinking,}}

\texttt{\textbf{then share it with the group.}}

\texttt{\textbf{Then all experts will go on to the next step, etc.}}

\texttt{\textbf{If any expert realises they're wrong at any point then they leave.}}

\texttt{\textbf{The question is...}}

\texttt{Based on the above information, who among the individuals would most benefit from receiving helpful information? Answer from the following options: A. Sophia B. Jayden C. Logan D. None of the above.}

\texttt{End your answer with 'Thus, the final answer is A (or B or C or D)'}

\texttt{Answer:
}

\paragraph{Self-Ask}

\texttt{You are a clever and helpful AI assistant. Your goal is to help human users around you.}

\texttt{One day you have the following observations that follow chronological order: }

\texttt{Lucas entered the cellar. Elizabeth entered the cellar. Ava entered the cellar. The pear is in the basket. Elizabeth exited the cellar. Lucas exited the cellar. Ava moved the pear to the suitcase. Ava exited the cellar. Ava dislikes the slippers. Elizabeth entered the study. Ava and Elizabeth plan to use the pear soon.}

\texttt{Based on the above information, who among the individuals would most benefit from receiving helpful information? Answer from the following options: A. Lucas B. Elizabeth C. Ava D. None of the above.
}

\texttt{\textbf{I will answer by first coming up and answering useful follow up questions and then reason slowly by considering all the follow up questions and answers, and finally come up with a final answer.}}

\texttt{\textbf{Format answer as follows:}}

\texttt{\textbf{Are follow up questions needed here: Yes.}}

\texttt{\textbf{Follow up: }}

\texttt{\textbf{Intermediate answer: }}

\texttt{\textbf{Follow up: }}

\texttt{\textbf{Intermediate answer:}}

\texttt{\textbf{Follow up: }}

\texttt{\textbf{Intermediate answer: }}

\texttt{\textbf{Let's reason to get a final answer by considering all above follow up questions and answers:}}

\texttt{\textbf{So the final answer is:}}

\texttt{End your answer with 'Thus, the final answer is A (or B or C or D)'}

\texttt{Answer:
}

\paragraph{FaR}

\texttt{You are a clever and helpful AI assistant. Your goal is to help human users around you.}

\texttt{One day you have the following observations that follow chronological order: }

\texttt{Jacob entered the bathroom. Emma entered the bathroom. The carrot is in the basket. Aiden entered the back yard. Emma exited the bathroom. Jacob moved the carrot to the pantry. Jacob and Emma plan to use the carrot soon.}

\texttt{Based on the above information, who among the individuals would most benefit from receiving helpful information? Answer from the following options: A. Jacob B. Emma C. Aiden D. None of the above.
}

\texttt{\textbf{I will first think about likely future events and identify potential challenges that each individual might be facing. Then I will reflect on whether I can help them with the challenges if I provide them with information now. Finally, I will choose a final answer based on the reasoning.}}

\texttt{\textbf{Format answer as follows:}}

\texttt{\textbf{\{}}

  \texttt{\textbf{"Character A's likely future actions":}}
  
  \texttt{\textbf{"Potential challenge 1":}}
  
  \texttt{\textbf{"Can I help with it now by providing information?":}}
  
  \texttt{\textbf{"Potential challenge 2":}}
  
  \texttt{\textbf{"Can I help with it now by providing information?":}}
  
  \texttt{\textbf{"Potential challenge 3":}}
  
  \texttt{\textbf{"Can I help with it now by providing information?":}}

  \texttt{\textbf{"Character B's likely future actions":}}
  
  \texttt{\textbf{"Potential challenge 1":}}
  
  \texttt{\textbf{"Can I help with it now by providing information?":}}
  
  \texttt{\textbf{"Potential challenge 2":}}
  
  \texttt{\textbf{"Can I help with it now by providing information?":}}
  
  \texttt{\textbf{"Potential challenge 3":}}
  
  \texttt{\textbf{"Can I help with it now by providing information?":}}

  \texttt{\textbf{"Character C's likely future actions":}}
  
  \texttt{\textbf{"Potential challenge 1":}}
  
  \texttt{\textbf{"Can I help with it now by providing information?":}}
  
  \texttt{\textbf{"Potential challenge 2":}}
  
  \texttt{\textbf{"Can I help with it now by providing information?":}}
  
  \texttt{\textbf{"Potential challenge 3":}}
  
  \texttt{\textbf{"Can I help with it now by providing information?":}}

  \texttt{\textbf{"final reasoning considering all steps above":}}
  
  \texttt{\textbf{"final answer":}}
  
\texttt{\textbf{\}}}
  
\texttt{\textbf{Answer:}}

\texttt{End your answer with 'Thus, the final answer is A (or B or C or D)'}

\texttt{Answer:
}

\section{Generalization Test Details}\label{appdx:generalization}
\subsection{Story Structure Tests}
Here we show a concrete example for each of the story structure generalization test from~\citet{sclar2023minding}.

\textbf{D1: Double Room False Belief Tests:}

Lucas entered the playroom. Alexander entered the playroom. The plum is in the box. 

The box is in the playroom. Alexander exited the playroom. Lucas moved the plum to the pantry. 

The pantry is in the playroom. Lucas exited the playroom. Alexander entered the porch.  

Lucas entered the porch.  The strawberry is in the bathtub.  The bathtub is in the porch.  

Lucas exited the porch.  Alexander moved the strawberry to the envelope.  The envelope is in the porch.  

Alexander exited the porch. Lucas and Alexander plan to use the plum soon.

Based on the above information, who among the individuals would most benefit from receiving helpful information? 

Answer from the following options: A. Alexander B. Lucas C. None of the above.

\textbf{D2: Three Active Characters Story:}

Lucas entered the playroom. Alexander entered the playroom. Jacob entered the playroom. 

The plum is in the box. The box is in the playroom. Alexander exited the playroom. 

Lucas moved the plum to the pantry. The pantry is in the playroom. Lucas exited the playroom.  

Jacob moved the plum to the bathtub.  The bathtub is in the playroom.  Jacob exited the playroom. 

Lucas, Alexander, and Jacob plan to use the plum soon.

Based on the above information, who among the individuals would most benefit from receiving helpful information? 

Answer from the following options: A. Lucas B. Alexander C. Jacob D. Lucas and Alexander E. None of the above.

\textbf{D3: Four Containers with Multiple Movements:}

Mia is in the patio. The slippers is in the crate. The crate is in the patio. 

Mia moved the slippers to the suitcase. The suitcase is in the patio. Isla entered the patio. 

Mia exited the patio. Isla moved the slippers to the bucket. The bucket is in the patio.  

Isla moved the slippers to the container.  The container is in the patio. Mia and Isla plan to use the slippers soon.

Based on the above information, who among the individuals would most benefit from receiving helpful information? 

Answer from the following options: A. Mia B. Isla C. None of the above.

\subsection{Faux-Pas Tests}
\citet{shapira-etal-2023-well} examines LLMs' performance on Faux-Pas stories and release an expert-curated set of Faux-Pas stories.
We show an example of the story here:
Kim helped her Mum make an apple pie for her uncle when he came to visit. She carried it out of the kitchen. 

Kim's uncle is sitting on the couch in the living room. His home has many boxes of pies, most boxes are for pineapple, no apple boxes. 
He is looking forward to Kim's baking. 
Kim showed her uncle the pie she made…

Based on the above information, who would you like to help by providing emotional support? 

Answer from the following options: A. Kim B. Kim's Mum C. Uncle Tom D. None of the above.

%% file: T4D_FaR.bbl
\begin{thebibliography}{41}
\providecommand{\natexlab}[1]{#1}
\providecommand{\url}[1]{\texttt{#1}}
\expandafter\ifx\csname urlstyle\endcsname\relax
  \providecommand{\doi}[1]{doi: #1}\else
  \providecommand{\doi}{doi: \begingroup \urlstyle{rm}\Url}\fi

\bibitem[Andreas(2022)]{andreas-2022-language}
Jacob Andreas.
\newblock Language models as agent models.
\newblock In \emph{Findings of the Association for Computational Linguistics:
  EMNLP 2022}, pp.\  5769--5779, Abu Dhabi, United Arab Emirates, December
  2022. Association for Computational Linguistics.
\newblock URL \url{https://aclanthology.org/2022.findings-emnlp.423}.

\bibitem[Anil et~al.(2023)Anil, Dai, Firat, Johnson, Lepikhin, Passos, Shakeri,
  Taropa, Bailey, Chen, et~al.]{anil2023palm}
Rohan Anil, Andrew~M Dai, Orhan Firat, Melvin Johnson, Dmitry Lepikhin,
  Alexandre Passos, Siamak Shakeri, Emanuel Taropa, Paige Bailey, Zhifeng Chen,
  et~al.
\newblock Palm 2 technical report.
\newblock \emph{arXiv preprint arXiv:2305.10403}, 2023.

\bibitem[Bara et~al.(2021)Bara, Sky, and Chai]{bara2021mindcraft}
Cristian-Paul Bara, CH-Wang Sky, and Joyce Chai.
\newblock Mindcraft: Theory of mind modeling for situated dialogue in
  collaborative tasks.
\newblock In \emph{Proceedings of the 2021 Conference on Empirical Methods in
  Natural Language Processing}, pp.\  1112--1125, 2021.

\bibitem[Baron-Cohen et~al.(1985)Baron-Cohen, Leslie, and Frith]{baron1985does}
Simon Baron-Cohen, Alan~M Leslie, and Uta Frith.
\newblock Does the autistic child have a “theory of mind”?
\newblock \emph{Cognition}, 21\penalty0 (1):\penalty0 37--46, 1985.

\bibitem[Baron-Cohen et~al.(1999)Baron-Cohen, O'riordan, Stone, Jones, and
  Plaisted]{baron1999recognition}
Simon Baron-Cohen, Michelle O'riordan, Valerie Stone, Rosie Jones, and Kate
  Plaisted.
\newblock Recognition of faux pas by normally developing children and children
  with asperger syndrome or high-functioning autism.
\newblock \emph{Journal of autism and developmental disorders}, 29:\penalty0
  407--418, 1999.

\bibitem[Besta et~al.(2023)Besta, Blach, Kubicek, Gerstenberger, Gianinazzi,
  Gajda, Lehmann, Podstawski, Niewiadomski, Nyczyk, et~al.]{got1}
Maciej Besta, Nils Blach, Ales Kubicek, Robert Gerstenberger, Lukas Gianinazzi,
  Joanna Gajda, Tomasz Lehmann, Michal Podstawski, Hubert Niewiadomski, Piotr
  Nyczyk, et~al.
\newblock Graph of thoughts: Solving elaborate problems with large language
  models.
\newblock \emph{arXiv preprint arXiv:2308.09687}, 2023.

\bibitem[Brown et~al.(2020)Brown, Mann, Ryder, Subbiah, Kaplan, Dhariwal,
  Neelakantan, Shyam, Sastry, Askell, et~al.]{brown2020language}
Tom~B Brown, Benjamin Mann, Nick Ryder, Melanie Subbiah, Jared Kaplan, Prafulla
  Dhariwal, Arvind Neelakantan, Pranav Shyam, Girish Sastry, Amanda Askell,
  et~al.
\newblock Language models are few-shot learners.
\newblock \emph{arXiv preprint arXiv:2005.14165}, 2020.

\bibitem[Fiske(1992)]{fiske1992thinking}
Susan~T Fiske.
\newblock Thinking is for doing: portraits of social cognition from
  daguerreotype to laserphoto.
\newblock \emph{Journal of personality and social psychology}, 63\penalty0
  (6):\penalty0 877, 1992.

\bibitem[Frith \& Frith(2003)Frith and Frith]{frith2003development}
Uta Frith and Christopher~D Frith.
\newblock Development and neurophysiology of mentalizing.
\newblock \emph{Philosophical Transactions of the Royal Society of London.
  Series B: Biological Sciences}, 358\penalty0 (1431):\penalty0 459--473, 2003.

\bibitem[Gur et~al.(2023)Gur, Furuta, Huang, Safdari, Matsuo, Eck, and
  Faust]{gur2023real}
Izzeddin Gur, Hiroki Furuta, Austin Huang, Mustafa Safdari, Yutaka Matsuo,
  Douglas Eck, and Aleksandra Faust.
\newblock A real-world webagent with planning, long context understanding, and
  program synthesis.
\newblock \emph{arXiv preprint arXiv:2307.12856}, 2023.

\bibitem[Hao et~al.(2023)Hao, Gu, Ma, Hong, Wang, Wang, and Hu]{rap}
Shibo Hao, Yi~Gu, Haodi Ma, Joshua~Jiahua Hong, Zhen Wang, Daisy~Zhe Wang, and
  Zhiting Hu.
\newblock Reasoning with language model is planning with world model.
\newblock \emph{arXiv preprint arXiv:2305.14992}, 2023.

\bibitem[Hart et~al.(1968)Hart, Nilsson, and Raphael]{hart1968formal}
Peter~E Hart, Nils~J Nilsson, and Bertram Raphael.
\newblock A formal basis for the heuristic determination of minimum cost paths.
\newblock \emph{IEEE transactions on Systems Science and Cybernetics},
  4\penalty0 (2):\penalty0 100--107, 1968.

\bibitem[Kojima et~al.(2022)Kojima, Gu, Reid, Matsuo, and
  Iwasawa]{kojima2022large}
Takeshi Kojima, Shixiang~Shane Gu, Machel Reid, Yutaka Matsuo, and Yusuke
  Iwasawa.
\newblock Large language models are zero-shot reasoners.
\newblock \emph{Advances in neural information processing systems},
  35:\penalty0 22199--22213, 2022.

\bibitem[Kosinski(2023)]{kosinski2023theory}
Michal Kosinski.
\newblock Theory of mind may have spontaneously emerged in large language
  models.
\newblock \emph{arXiv preprint arXiv:2302.02083}, 2023.

\bibitem[Le et~al.(2019)Le, Boureau, and Nickel]{le2019revisiting}
Matthew Le, Y-Lan Boureau, and Maximilian Nickel.
\newblock Revisiting the evaluation of theory of mind through question
  answering.
\newblock In \emph{Proceedings of the 2019 Conference on Empirical Methods in
  Natural Language Processing and the 9th International Joint Conference on
  Natural Language Processing (EMNLP-IJCNLP)}, pp.\  5872--5877, 2019.

\bibitem[Mahowald et~al.(2023)Mahowald, Ivanova, Blank, Kanwisher, Tenenbaum,
  and Fedorenko]{mahowald2023dissociating}
Kyle Mahowald, Anna~A Ivanova, Idan~A Blank, Nancy Kanwisher, Joshua~B
  Tenenbaum, and Evelina Fedorenko.
\newblock Dissociating language and thought in large language models: a
  cognitive perspective.
\newblock \emph{arXiv preprint arXiv:2301.06627}, 2023.

\bibitem[Mishra et~al.(2021)Mishra, Khashabi, Baral, Choi, and
  Hajishirzi]{mishra_reframing_2021}
Swaroop Mishra, Daniel Khashabi, Chitta Baral, Yejin Choi, and Hannaneh
  Hajishirzi.
\newblock \href{http://arxiv.org/abs/2109.07830}{Reframing {Instructional}
  {Prompts} to {GPTk}'s {Language}}.
\newblock \emph{arXiv preprint arXiv:2109.07830}, 2021.

\bibitem[Nematzadeh et~al.(2018)Nematzadeh, Burns, Grant, Gopnik, and
  Griffiths]{nematzadeh2018evaluating}
Aida Nematzadeh, Kaylee Burns, Erin Grant, Alison Gopnik, and Tom Griffiths.
\newblock Evaluating theory of mind in question answering.
\newblock In \emph{Proceedings of the 2018 Conference on Empirical Methods in
  Natural Language Processing}, pp.\  2392--2400, 2018.

\bibitem[OpenAI(2022)]{chatgpt}
OpenAI.
\newblock Chatgpt: Optimizing language models for dialogue, 2022.
\newblock URL \url{https://openai.com/blog/chatgpt/}.

\bibitem[OpenAI(2023)]{openai2023gpt}
R~OpenAI.
\newblock Gpt-4 technical report.
\newblock \emph{arXiv}, pp.\  2303--08774, 2023.

\bibitem[Park et~al.(2023)Park, O'Brien, Cai, Morris, Liang, and
  Bernstein]{park2023generative}
Joon~Sung Park, Joseph~C O'Brien, Carrie~J Cai, Meredith~Ringel Morris, Percy
  Liang, and Michael~S Bernstein.
\newblock Generative agents: Interactive simulacra of human behavior.
\newblock \emph{arXiv preprint arXiv:2304.03442}, 2023.

\bibitem[Perner et~al.(1987)Perner, Leekam, and Wimmer]{perner1987three}
Josef Perner, Susan~R Leekam, and Heinz Wimmer.
\newblock Three-year-olds' difficulty with false belief: The case for a
  conceptual deficit.
\newblock \emph{British journal of developmental psychology}, 5\penalty0
  (2):\penalty0 125--137, 1987.

\bibitem[Premack \& Woodruff(1978)Premack and Woodruff]{premack1978does}
David Premack and Guy Woodruff.
\newblock Does the chimpanzee have a theory of mind?
\newblock \emph{Behavioral and brain sciences}, 1\penalty0 (4):\penalty0
  515--526, 1978.

\bibitem[Press et~al.(2022)Press, Zhang, Min, Schmidt, Smith, and
  Lewis]{press2022measuring}
Ofir Press, Muru Zhang, Sewon Min, Ludwig Schmidt, Noah~A Smith, and Mike
  Lewis.
\newblock Measuring and narrowing the compositionality gap in language models.
\newblock \emph{arXiv preprint arXiv:2210.03350}, 2022.

\bibitem[Sap et~al.(2019)Sap, Rashkin, Chen, Le~Bras, and Choi]{sap2019social}
Maarten Sap, Hannah Rashkin, Derek Chen, Ronan Le~Bras, and Yejin Choi.
\newblock Social {IQ}a: Commonsense reasoning about social interactions.
\newblock In \emph{Proceedings of the 2019 Conference on Empirical Methods in
  Natural Language Processing and the 9th International Joint Conference on
  Natural Language Processing (EMNLP-IJCNLP)}, pp.\  4463--4473, Hong Kong,
  China, 2019. Association for Computational Linguistics.
\newblock \doi{10.18653/v1/D19-1454}.
\newblock URL \url{https://aclanthology.org/D19-1454}.

\bibitem[Sap et~al.(2022)Sap, Le~Bras, Fried, and Choi]{sap-etal-2022-neural}
Maarten Sap, Ronan Le~Bras, Daniel Fried, and Yejin Choi.
\newblock Neural theory-of-mind? on the limits of social intelligence in large
  {LM}s.
\newblock In \emph{Proceedings of the 2022 Conference on Empirical Methods in
  Natural Language Processing}, pp.\  3762--3780, Abu Dhabi, United Arab
  Emirates, December 2022. Association for Computational Linguistics.
\newblock URL \url{https://aclanthology.org/2022.emnlp-main.248}.

\bibitem[Schick et~al.(2023)Schick, Dwivedi-Yu, Dess{\`\i}, Raileanu, Lomeli,
  Zettlemoyer, Cancedda, and Scialom]{schick2023toolformer}
Timo Schick, Jane Dwivedi-Yu, Roberto Dess{\`\i}, Roberta Raileanu, Maria
  Lomeli, Luke Zettlemoyer, Nicola Cancedda, and Thomas Scialom.
\newblock Toolformer: Language models can teach themselves to use tools.
\newblock \emph{arXiv preprint arXiv:2302.04761}, 2023.

\bibitem[Sclar et~al.(2023)Sclar, Kumar, West, Suhr, Choi, and
  Tsvetkov]{sclar2023minding}
Melanie Sclar, Sachin Kumar, Peter West, Alane Suhr, Yejin Choi, and Yulia
  Tsvetkov.
\newblock Minding language models'(lack of) theory of mind: A plug-and-play
  multi-character belief tracker.
\newblock \emph{arXiv preprint arXiv:2306.00924}, 2023.

\bibitem[Shapira et~al.(2023{\natexlab{a}})Shapira, Levy, Alavi, Zhou, Choi,
  Goldberg, Sap, and Shwartz]{shapira2023clever}
Natalie Shapira, Mosh Levy, Seyed~Hossein Alavi, Xuhui Zhou, Yejin Choi, Yoav
  Goldberg, Maarten Sap, and Vered Shwartz.
\newblock Clever hans or neural theory of mind? stress testing social reasoning
  in large language models.
\newblock \emph{arXiv preprint arXiv:2305.14763}, 2023{\natexlab{a}}.

\bibitem[Shapira et~al.(2023{\natexlab{b}})Shapira, Zwirn, and
  Goldberg]{shapira-etal-2023-well}
Natalie Shapira, Guy Zwirn, and Yoav Goldberg.
\newblock How well do large language models perform on faux pas tests?
\newblock In \emph{Findings of the Association for Computational Linguistics:
  ACL 2023}, pp.\  10438--10451, Toronto, Canada, July 2023{\natexlab{b}}.
  Association for Computational Linguistics.
\newblock \doi{10.18653/v1/2023.findings-acl.663}.
\newblock URL \url{https://aclanthology.org/2023.findings-acl.663}.

\bibitem[Shapira et~al.(2022)Shapira, Pasunuru, Bansal, Dagan, and
  Amsterdamer]{shapira-etal-2022-interactive}
Ori Shapira, Ramakanth Pasunuru, Mohit Bansal, Ido Dagan, and Yael Amsterdamer.
\newblock Interactive query-assisted summarization via deep reinforcement
  learning.
\newblock In \emph{Proceedings of the 2022 Conference of the North American
  Chapter of the Association for Computational Linguistics: Human Language
  Technologies}, pp.\  2551--2568, Seattle, United States, July 2022.
  Association for Computational Linguistics.
\newblock \doi{10.18653/v1/2022.naacl-main.184}.
\newblock URL \url{https://aclanthology.org/2022.naacl-main.184}.

\bibitem[Strathman et~al.(1994)Strathman, Gleicher, Boninger, and
  Edwards]{strathman1994consideration}
Alan Strathman, Faith Gleicher, David~S Boninger, and C~Scott Edwards.
\newblock The consideration of future consequences: Weighing immediate and
  distant outcomes of behavior.
\newblock \emph{Journal of personality and social psychology}, 66\penalty0
  (4):\penalty0 742, 1994.

\bibitem[Trott et~al.(2023)Trott, Jones, Chang, Michaelov, and
  Bergen]{trott2023large}
Sean Trott, Cameron Jones, Tyler Chang, James Michaelov, and Benjamin Bergen.
\newblock Do large language models know what humans know?
\newblock \emph{Cognitive Science}, 47\penalty0 (7):\penalty0 e13309, 2023.

\bibitem[Ullman(2023)]{ullman2023large}
Tomer Ullman.
\newblock Large language models fail on trivial alterations to theory-of-mind
  tasks.
\newblock \emph{arXiv preprint arXiv:2302.08399}, 2023.

\bibitem[Wei et~al.(2022)Wei, Wang, Schuurmans, Bosma, Xia, Chi, Le, Zhou,
  et~al.]{wei2022chain}
Jason Wei, Xuezhi Wang, Dale Schuurmans, Maarten Bosma, Fei Xia, Ed~Chi, Quoc~V
  Le, Denny Zhou, et~al.
\newblock Chain-of-thought prompting elicits reasoning in large language
  models.
\newblock \emph{Advances in Neural Information Processing Systems},
  35:\penalty0 24824--24837, 2022.

\bibitem[Wimmer \& Perner(1983)Wimmer and Perner]{wimmer1983beliefs}
Heinz Wimmer and Josef Perner.
\newblock Beliefs about beliefs: Representation and constraining function of
  wrong beliefs in young children's understanding of deception.
\newblock \emph{Cognition}, 13\penalty0 (1):\penalty0 103--128, 1983.

\bibitem[Yao et~al.(2023{\natexlab{a}})Yao, Yu, Zhao, Shafran, Griffiths, Cao,
  and Narasimhan]{yao2023tree}
Shunyu Yao, Dian Yu, Jeffrey Zhao, Izhak Shafran, Thomas~L Griffiths, Yuan Cao,
  and Karthik Narasimhan.
\newblock Tree of thoughts: Deliberate problem solving with large language
  models.
\newblock \emph{arXiv preprint arXiv:2305.10601}, 2023{\natexlab{a}}.

\bibitem[Yao et~al.(2023{\natexlab{b}})Yao, Zhao, Yu, Du, Shafran, Narasimhan,
  and Cao]{yao2023react}
Shunyu Yao, Jeffrey Zhao, Dian Yu, Nan Du, Izhak Shafran, Karthik Narasimhan,
  and Yuan Cao.
\newblock {ReAct}: Synergizing reasoning and acting in language models.
\newblock In \emph{International Conference on Learning Representations
  (ICLR)}, 2023{\natexlab{b}}.

\bibitem[Yao et~al.(2023{\natexlab{c}})Yao, Li, and Zhao]{got2}
Yao Yao, Zuchao Li, and Hai Zhao.
\newblock Beyond chain-of-thought, effective graph-of-thought reasoning in
  large language models.
\newblock \emph{arXiv preprint arXiv:2305.16582}, 2023{\natexlab{c}}.

\bibitem[Zhou et~al.(2022)Zhou, Sch{\"a}rli, Hou, Wei, Scales, Wang,
  Schuurmans, Bousquet, Le, and Chi]{zhou2022least}
Denny Zhou, Nathanael Sch{\"a}rli, Le~Hou, Jason Wei, Nathan Scales, Xuezhi
  Wang, Dale Schuurmans, Olivier Bousquet, Quoc Le, and Ed~Chi.
\newblock \href{https://arxiv.org/abs/2205.10625}{Least-to-Most Prompting
  Enables Complex Reasoning in Large Language Models}.
\newblock \emph{arXiv preprint arXiv:2205.10625}, 2022.

\bibitem[Zhou et~al.(2023)Zhou, Zhu, Hu, Pujara, Ren, Callison-Burch, Choi, and
  Ammanabrolu]{zhou-etal-2023-cast}
Pei Zhou, Andrew Zhu, Jennifer Hu, Jay Pujara, Xiang Ren, Chris Callison-Burch,
  Yejin Choi, and Prithviraj Ammanabrolu.
\newblock {I} cast detect thoughts: Learning to converse and guide with intents
  and theory-of-mind in dungeons and dragons.
\newblock In \emph{Proceedings of the 61st Annual Meeting of the Association
  for Computational Linguistics (Volume 1: Long Papers)}, pp.\  11136--11155,
  Toronto, Canada, July 2023. Association for Computational Linguistics.
\newblock \doi{10.18653/v1/2023.acl-long.624}.
\newblock URL \url{https://aclanthology.org/2023.acl-long.624}.

\end{thebibliography}
